\pdfoutput=1
\documentclass[11pt]{article}

\usepackage[]{EMNLP2022}

\usepackage{times}
\usepackage{latexsym}

\usepackage[T1]{fontenc}
\usepackage[utf8]{inputenc}

\usepackage{microtype}

\usepackage{inconsolata}

\title{Fine-tuned Language Models are Continual Learners}

\author{Thomas Scialom$^1$\thanks{~~Both Authors Contributed Equally}~~~~~Tuhin Chakrabarty$^{2*}$~~~~~Smaranda Muresan $^{2}$ \\
 $^1$Meta AI\\  
 $^2$Department of Computer Science, Columbia University\\
 {\tt\small tscialom@fb.com, tuhin.chakr@cs.columbia.edu, smara@cs.columbia.edu} \\
}

\usepackage{times,latexsym}
\usepackage{url}
\usepackage[T1]{fontenc}
\usepackage{multirow}
\usepackage{multicol}

\usepackage{enumitem} 
\usepackage{booktabs}
\usepackage{comment}
\usepackage{graphicx}
\usepackage{subfigure}

\begin{document}
\maketitle
\begin{abstract}

Recent work on large language models relies on the intuition that most natural language processing tasks can be described via natural language instructions and that models trained on these instructions show strong zero-shot performance on several standard datasets. However, these models even though impressive still perform poorly on a wide range of tasks outside of their respective training and evaluation sets. To address this limitation, we argue that a model should be able to keep extending its knowledge and abilities, without forgetting previous skills. In spite of the limited success of Continual Learning we show that \emph{Fine-tuned Language Models can be continual learners}. We empirically investigate the reason for this success and conclude that Continual Learning emerges from self-supervision pre-training. Our resulting model Continual-T0 (CT0) is able to learn 8 new diverse language generation tasks, while still maintaining good performance on previous tasks, spanning in total 70 datasets. Finally, we show that CT0 is able to combine instructions in ways it was never trained for, demonstrating some level of instruction compositionality.\footnote{Our code is publicly available at \url{https://github.com/ThomasScialom/T0_continual_learning}. }

\end{abstract}

%\iftaclpubformat

\section{Introduction}
\label{sec:intro}

Recent work has shown that large language models have the ability to perform zero-shot and few-shot learning reasonably well \cite{brown2020language,rae2021scaling,smith2022using}. A particularly successful line of work relies on the intuition that most natural language processing tasks can be described via natural language instructions. For example, a summarization task can be reformatted as a response to a natural language input as shown in Table \ref{table:t0}. \newcite{sanh2021multitask} and \newcite{wei2021finetuned} have released T0 and FLAN respectively and shown that fine-tuning models on a massive mixture of NLP datasets expressed via such natural language instructions (i.e., instruction tuning), improves the zero-shot performance of large language models. FLAN is extremely large in size (137B) and is not publicly available limiting its further use and reproducibility. Conversely T0 \cite{sanh2021multitask} is publicly available and orders of magnitude smaller and hence we resort to working with T0.

\begin{table}[]
\centering
\small
\renewcommand{\arraystretch}{1.15}
\begin{tabular}{|l|}
\hline
\begin{tabular}[c]{@{}l@{}}The picture appeared on the wall of a\\ Poundland store on Whymark Avenue {[}...{]} \textbf{How}\\ \textbf{would you rephrase that in a few words?}\end{tabular} \\ \hline
\begin{tabular}[c]{@{}l@{}}Graffiti artist Banksy is believed to be\\ behind {[}....{]}\end{tabular}                                                          \\ \hline
\end{tabular}
\vspace{-0.1cm}
\caption{\label{table:t0}An instance from T0 training set \cite{sanh2021multitask} where a summarization task is reformatted as a natural language response to a natural language input.}
\vspace{-0.5cm}
\end{table}

However impressive, these models are still limited to simple instructions and mainly Natural Language Understanding (NLU) tasks. 
These models perform poorly on a wide range of tasks that differ from their respective evaluation sets. To improve their ability on new and diverse tasks, one needs to fine-tune these models again. However, one key problem associated with fine-tuning is \textit{catastrophic forgetting} \cite{french1999catastrophic}. So, how can we extend these models knowledge and abilities, without suffering from catastrophic forgetting?

In this paper, we study Continual Learning of large language models \emph{fine-tuned on natural language instructions} and investigate their ability to adapt to diverse tasks, while avoiding catastrophic forgetting on the older tasks. For this purpose,  we propose Continual-T0 (CT0), a T0 model that uses Continual Learning with rehearsal \cite{shin2017continual}, i.e. using a memory buffer containing a small portion of previous data replayed during training (Section \ref{sec:ContinualLearningRehearsal}).
We start from T0, a model trained jointly on 50 datasets, resulting in a good zero-shot performance on 12 completely different datasets. We are then able to teach progressively 8 new diverse tasks, while maintaining almost 100\% of the initial performance on all the previous datasets. 
This result is obtained by using only 1\% of data for memory buffer. Notably, we also maintain the performance for the T0 zero-shot evaluation datasets, even though no rehearsal for those was done, the first of its kind setup for CL (Section \ref{sec:results}). %SM-f 

Our final model, Continual-T0 (CT0) in addition to performing as well as T0 on all the different T0 datasets, can also understand instructions about the 
newly introduced tasks focused on language generation problems such as writing a haiku, generating empathetic responses in a dialogue, simplifying text, generating a headline with decoding constraints, generating natural language explanations for Natural Language Inferece (NLI) tasks, generating a Tweet on a given topic in the style of a given author, or question answering for new domain/concepts such as COVID-19.

We also conduct an extensive analysis and show that our newly learned instructions can be composed in ways never seen during training, leading to better generalization (Section \ref{sec:compositionality}).
Given the surprising performance of a simple continual learning strategy, we empirically investigate the reason for this success. Why transformer models like T0 are continual learners? Is it because of their multi-task nature or the instruction tuning paradigm? Or does the large scale parameterization of language models contribute to this success? Our experimental analysis show that the easy adaptability and continual learning capabilities actually emerge from pre-training and not the above, \textbf{including scale} (Table \ref{tab:appendix_t5}, Section \ref{whyllm}).

\section{Related Work}
\label{sec:related_work}

\paragraph{Continual Learning}

Current models are limited in continuously learning without forgetting any previously acquired knowledge and abilities. Research in this direction has investigated various strategies such as External Memory, Constraints and Model Plasticity \cite{parisi2019continual}. External Memory methods often simply use rehearsal with a replay during training \cite{rebuffi2017icarl}. \citet{de2019episodic} also proposed local fine-tuning at inference time, leveraging examples similar to the considered input.

Through the lens of NLP tasks, \citet{biesialska-etal-2020-continual} look at the problem of CL and discuss major challenges involved. \citet{jin2021lifelong} show CL algorithms are effective for knowledge preservation. Their study also infer that continual pretraining improves temporal generalization. \cite{douillard2021dytox} proposed a a dynamic expansion of special tokens with a transformer architecture. \citet{mi-etal-2020-continual} and \citet{madotto-etal-2021-continual} perform CL for task oriented dialog systems by using replay based strategy. \citet{cao-etal-2021-continual} propose a new CL framework for NMT models, while \citet{ke-etal-2021-adapting} proposes a novel capsule network based model called B-CL (Bert based CL) for  sentiment classification tasks. \citet{jin-etal-2020-visually} show how existing CL algorithms fail at learning compositional phrases. \citet{lin-etal-2022-continual} propose a benchmark and highlight key challenges for continual model refinement in Out-of-Distribution data streams. More recently,  \citet{sun2019lamol} propose a lifelong learning method LAMOL that is capable of continually learning new tasks by replaying pseudo-samples of previous tasks that require no extra memory or model capacity. To the best of our knowledge, LAMOL corresponds to the state-of-the-art for CL in NLP. Most similar to our work is that of \citet{yin-etal-2022-contintin} who also study continual learning from task instructions based on the NATURAL-INSTRUCTION benchmark \cite{mishra2021cross}.Finally instead of limiting to vision-only and language-only tasks \citet{srinivasan2022climb} study the challenge of learning multimodal tasks in a CL setting, and systematically evaluate how upstream continual learning can rapidly generalize to new multimodal and unimodal tasks 

Most of the aforementioned works fall into the 2 scenarios differentiated by \citet{lomonaco2017core50}: 1) learning new data of known classes (online learning), and 2) learning new classes (class-incremental learning). Thus, the study are often limited to a narrow domain, or a specific task. In our work, we propose to address Continual Learning more broadly: learning a diverse set of new tasks different from the ones used for training. For this, we leverage the idea of instruction tuning \cite{wei2021finetuned,sanh2021multitask}, that enables us to frame any NLP task as a response to a natural language input and use rehearsal as a mechanism to avoid catastrophic forgetting \cite{shin2017continual}.

\section{Continual Learning for Fine-tuned Language Models}
\label{sec:ContinualLearningRehearsal}

\subsection{Continual Learning via Rehearsal (CLR)}

Our objective is to maintain the model's existing learned skills, while progressively learning more tasks. To prevent the model from catastrophic forgetting, we rely on an external memory module, storing a subset of  previous training data \cite{shin2017continual}. %SM-f I added training before data? 
We define the tasks to be learned as a task sequence $T = (T_1, T_2, · · · , T_N )$ of $N$ tasks. $D_i$ is the corresponding dataset for task $T_i$. Formally, the training data augmented with \textbf{r}ehearsal $D_{i}^r$ is defined as:
 \vspace{-0.3cm}
\begin{equation}
    \vspace{-0.2cm}
    \label{equ:rehearsal}
    D_{i}^r = D_{i} \bigcup \sum_{j=1}^{i-1}(r D_{j})
\end{equation}
where r is the rehearsal hyper-parameter that controls the percentage of examples sampled from previous tasks $T_1, ... T_{i-1}$. We note that $r=0$ corresponds to no memory, and $r=1$ is equivalent to a multi-task setup using all the previous examples.

\subsection{Continual-T0 (CT0)}
\label{sec:implem_detail}
For all our experiments, we instantiate our model with the T0 model \cite{sanh2021multitask}. T0 is a T5 model \cite{raffel2020exploring} fine-tuned in a multitask setting on 50 datasets, where the natural language instructions corresponding to individual tasks are used as the input. The set of these 50 tasks corresponds therefore to $T_1$ in \ref{equ:rehearsal}. This massive instruction tuning allows the model to perform well in a zero-shot setup, by leveraging the information presents only in the instructions. Our initial model is T0\_3B, the T0 version with (only) 3 Billions parameters for all our experiments. We used the same hyper-parameters as the ones reported in \citet{sanh2021multitask}\footnote{See more details at \url{https://huggingface.co/bigscience/T0pp}}. The only new hyper-parameter introduced in our paper is the \emph{rehearsal proportion} $r$. We explore $r \in [0, 0.25\%, 1\%]$ as reported in our first set of results (see Section \ref{sec:ContinualLearningRehearsal}).

For each of T0 training tasks,we consider 100,000 examples for training, such that 1\% rehearsal corresponds to 1,000 examples that will be used as the memory buffer for rehearsal. Thus, for datasets with fewer training examples, we upsample them and conversely for largest datasets like Gigaword or Simplification, we limit to 100,000 examples. Note that here, \textbf{while we used rehearsal for the \emph{training} data of T0 training tasks, we never used any data from T0 zeroshot tasks, so it remains completely zero-shot}. It is important to highlight that rehearsal is the standard for CL, and a zero-shot set up with no rehearsal has never been explored yet to the best of our knowledge.

\begin{table*}[!ht]
\centering
\small
\renewcommand{\arraystretch}{1.15}
\begin{tabular}{|p{1.8cm}|p{1.35cm}|ll|}
\hline
\multirow{2}{*}{\begin{tabular}[c]{@{}l@{}}Text\\ Smpfl(Simp)\end{tabular}}                                                 & Instruction & \multicolumn{2}{l|}{{\color{blue}Make this text simpler:} "A Georgian inscription around the drum attests his name."}                                                                                                                                                                                                                                                                                                                                                                                                                                                                                                         \\ \cline{2-4} 
                                                                                & Output & \multicolumn{2}{l|}{\textit{A Georgian writing on the drum is his name.}}                                                                                                                                                                                                                                                                                                                                                                                                                                                                                                                                                \\ \hline\hline
\multirow{3}{*}{\begin{tabular}[c]{@{}l@{}}Headline\\ Generation \\(HGen)\end{tabular}} & Instruction & \multicolumn{2}{l|}{\begin{tabular}[c]{@{}l@{}}{\color{blue}Make a title for this article that begins with "protesters":} police arrested five\\ anti-nuclear  protesters thursday after they sought to disrupt loading of a french \\antarctic research and supply vessel , a spokesman for the protesters said .\end{tabular}}                                                                                                                                                                                                                                                                                              \\ \cline{2-4} 
                                                                                & Output & \multicolumn{2}{l|}{\textit{protesters target french research ship}}                                                                                                                                                                                                                                                                                                                                                                                                                                                                                                                                                     \\ \hline\hline
\multirow{2}{*}{\begin{tabular}[c]{@{}l@{}}Haiku\\ Gen (Haiku)\end{tabular}}                                                         & Instruction & \multicolumn{2}{l|}{{\color{blue}Generate a haiku about} 'Seagulls crying high'}                                                                                                                                                                                                                                                                                                                                                                                                                                                                                                                                              \\ \cline{2-4} 
                                                                                & Output & \multicolumn{2}{l|}{\textit{Seagulls crying high /  the air smelling of sea salt /  Or is it my tears?}}                                                                                                                                                                                                                                                                                                                                                                                                                                                                                                                   \\ \hline\hline
\multirow{2}{*}{\begin{tabular}[c]{@{}l@{}}Covid QA\\ (CQA)\end{tabular}}                                                        & Instruction & \multicolumn{2}{l|}{\begin{tabular}[c]{@{}l@{}}{\color{blue}In the context of the COVID pandemic},who is at greater risk of dying  \\ from COVID19?\end{tabular}}                                                                                                                                                                                                                                                                                                                                                                                                                                                    \\ \cline{2-4} 
                                                                                & Output & \multicolumn{2}{l|}{\begin{tabular}[c]{@{}l@{}}\textit{patients with underlying medical conditions and the elderly}\end{tabular}}                                                                                                                                                                                                                                                                                                                                                                                                                                                 \\ \hline\hline
\multirow{3}{*}{\begin{tabular}[c]{@{}l@{}}Inquisitive\\ Question \\Gen(InqQG)\end{tabular}}                                                           & Instruction & \multicolumn{2}{l|}{\begin{tabular}[c]{@{}l@{}}{\color{blue}Given the following text, write the possible curious question it answers: }"Positrons \\do not travel backwards in time. Positron-electron annihilation ....... So, we know \\they collide frequently enough to light up the galaxy in that part of the spectrum\end{tabular}} \\ \cline{2-4} 
                                                                                & Output & \multicolumn{2}{l|}{\textit{How often do electrons and positrons collide in nature?}}                                                                                                                                                                                                                                                                                                                                                                                                                                                                                                                                 \\ \hline\hline
\multirow{4}{*}{\begin{tabular}[c]{@{}l@{}}Empathetic\\ Dialog\\ Generation \\(EmDg)\end{tabular}}    & Instruction & \multicolumn{2}{l|}{\begin{tabular}[c]{@{}l@{}}{\color{blue}The associated emotion is }"disappointed" {\color{blue}and the input prompt is} "Had to cancel our \\family vacation coming up next month. My husband's work said he couldn't go \\after they already approved the time off.". {\color{blue}Now what would be your response, given} \\{\color{blue}the following dialogue context:===} - I had to cancel our family vacation coming \\up next month.\end{tabular}}                                                                                                                                                                                          \\ \cline{2-4} 
                                                                                & Output & \multicolumn{2}{l|}{\textit{I am really sorry to hear that. I hope everything is alright.}}                                                                                                                                                                                                                                                                                                                                                                                                                                                                                                                              \\ \hline\hline
\multirow{3}{*}{\begin{tabular}[c]{@{}l@{}}Explanation\\ Generation \\(Exp)\end{tabular}}                                                          & Instruction & \multicolumn{2}{l|}{\begin{tabular}[c]{@{}l@{}}{\color{blue}Explain why the two following sentences are contradicting each other: }"Sentence 1: \\A statue at a museum that no seems to be looking at."; Sentence 2: "Tons of \\people are gathered around the statue."\end{tabular}}                                                                                                                                                                                                                                                                                                                                                                     \\ \cline{2-4} 
                                                                                & Output & \multicolumn{2}{l|}{\begin{tabular}[c]{@{}l@{}} \textit{If tons of people are gathered around the statue, it is not possible that no one seems to }\\ \textit{be looking at it.}\end{tabular}}                                                                                                                                                                                                                                                                                                                                                                                                                                           \\ \hline\hline
\multirow{3}{*}{\begin{tabular}[c]{@{}l@{}}Twitter\\ Stylometry\\(TwSt)\end{tabular}}   & Instruction & \multicolumn{2}{l|}{{\color{blue}Write a tweet about} \#WelcomeToNewYork, {\color{blue}in the style of} taylorswift13}                                                                                                                                                                                                                                                                                                                                                                                                                                                                                                                      \\ \cline{2-4} 
                                                                                & Output & \multicolumn{2}{l|}{\begin{tabular}[c]{@{}l@{}}\textit{GUYS. \#WelcomeToNewYork will be up on iTunes any minute now. }\\ \textit{This is not a drill!! GO GO GO}\end{tabular}}                                                                                                                                                                                                                                                                                                                                                                                                                                                    \\ \hline
\end{tabular}
\caption{\label{prompts}Example Instructions with their respective ground-truth for 8 new tasks learned continually from T0.}
\label{tab:task_prompt_illustration}
\end{table*}

\subsection{Tasks}
\label{sec:tasks}

%In this section,
We briefly describe all the tasks $T$ used to progressively train and evaluate our model (a more complete description is also given %provided 
in Appendix \ref{tasks}).  

\paragraph{T0 Tasks.} As detailed in Section \ref{sec:intro}, we instantiate our model with T0 weights. T0 is trained in a multi-task setting on a collection of 50 datasets spanning from QA, Classification to Summarization. We refer to this set of 50 datasets as \emph{T0 train (T0tr)}. To evaluate the true zero-shot performance for T0, the authors evaluated it on a set of 12 datasets corresponding to 4 tasks different from  \emph{T0 train}: Natural Language Inference, Co-reference resolution, Word sense disambiguation and Sentence completion. We refer to this set as \emph{T0 zero-shot (T0zs)}.

\paragraph{New Tasks.} To extend T0 capabilities and benchmark its performance in our continual learning setup,we introduce 8 new tasks focused on language generation, unlike the existing T0 evaluation tasks and majority of the T0 training tasks (except summarization). These tasks include: 1) \textbf{Text Simplification (Simpl)} with the goal of paraphrasing a given text using simple language, where we train our model on WikiAuto \citet{acl/JiangMLZX20} and evaluate it on the WikiAuto and  ASSET datasets \cite{alva-manchego-etal-2020-asset};  2) \textbf{Headline Generation with Constraint (HGen)}, where given a news article D and an input keyword X, the goal is to generate a headline that contains the keyword at the beginning, at the end or anywhere (see Table \ref{tab:task_prompt_illustration} for a sample instruction to generate a headline containing the keyword at the beginning).
To create the training data, we simply leverage the gold-reference to select the keyword X, such that our model is trained with consistent and plausible instructions; 3) \textbf{Haiku Generation (Haiku)}, where the task is to generate a Haiku --- a type of short form poetry originally from Japan --- given a topic (see Table \ref{tab:task_prompt_illustration} for a sample instruction). 
We train on pairs (Haiku, title) from Reddit and generate Haikus for novel topics at inference time; 4) \textbf{Covid QA (CQA)} \cite{moller-etal-2020-covid}, a Question answering task focusing on COVID-19. Because T0 has been extensively trained on a QA dataset, CovidQA in its original format simply requires domain transfer. To make the task more challenging, we propose to provide only the question as an input, now framing the task as ``learn the answer by heart'' in an encyclopedia style task. This way the task framing can be seen as a new strategy to incorporating knowledge and preventing the model from concept drift. 

5) \textbf{Inquisitive Question Generation (InqQG)} where we train our model on the ELI5 dataset \cite{fan-etal-2019-eli5} to generate questions that typically require long form answers;  
6) \textbf{Empathetic Dialogue Generation (EmDg)}, where we generate a response to a conversational context grounded in emotional situations using the Empathetic Dialogue data \citet{rashkin-etal-2019-towards}; 7) \textbf{Explanation Generation (Exp)} where we train a model on the eSNLI  \cite{NEURIPS2018_4c7a167b} benchmark to generate natural language explanations given a premise, hypothesis and a label (entail, contradict, neutral); 8) \textbf{Twitter Stylometry (TwSt)}, where we generate a relevant tweet given a hashtag and the tweet's author by fine-tuning on the data consisting of tweets from the top 20 most followed users in Twitter released by \citet{tareaf2017r}.
We illustrate the 8 new tasks with their instructions in Table \ref{tab:task_prompt_illustration}. A complete detailed description for all the 8 tasks with train, validation splits is available in the Appendix \ref{section:newtask}. 

\subsection{Automatic Metrics}
We report the accuracy for T0 zero-shot tasks, and standard metrics for NLG like BLEU\cite{papineni-etal:2002:Bleu} and SARI\cite{xu-etal-2016-optimizing} for Simplification, ROUGE \cite{lin2004rouge} for Headline Generation, or BERTScore \cite{Zhang-etal:2020:bertscore} \footnote{We use the BERTScore version based off deberta-mnli}for open-domain NLG tasks as it has been found to correlate well with human judgements.

We also designed customized metrics for some of the tasks.\footnote{All those metrics implementations are available in the publicly released code.} For instance, to evaluate Twitter Stylometry where the task is to generate a tweet in the style of the author, we trained a Ridge Classifier to predict the author given the evaluated tweet. For Haiku generation, we know that in general, a Haiku contains only 17 syllables, broken up into three lines. We therefore create a metric to reflect the task structure that integrates i) the differences in syllables and number of lines between the gold and generated haiku, ii) the BLEU score between gold and predicted, and  iii) the presence of the topic in the generated haiku. We report all the details for the metrics in the Appendix \ref{AutoNLG}.  

\section{Results}
\label{sec:results}

\begin{figure}[!ht]
    \includegraphics[width = 0.55\textwidth]{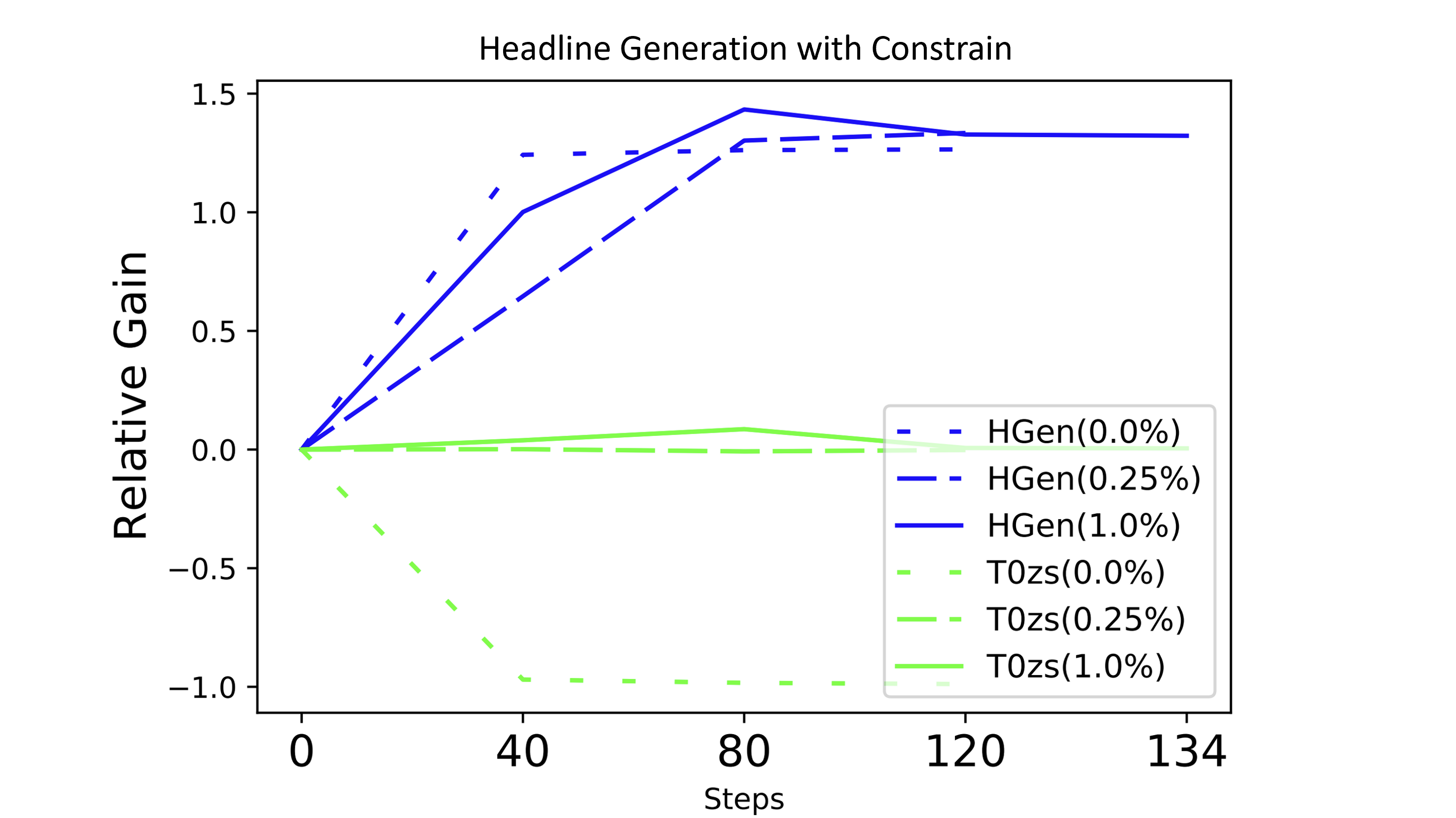}
    \includegraphics[width = 0.52\textwidth]{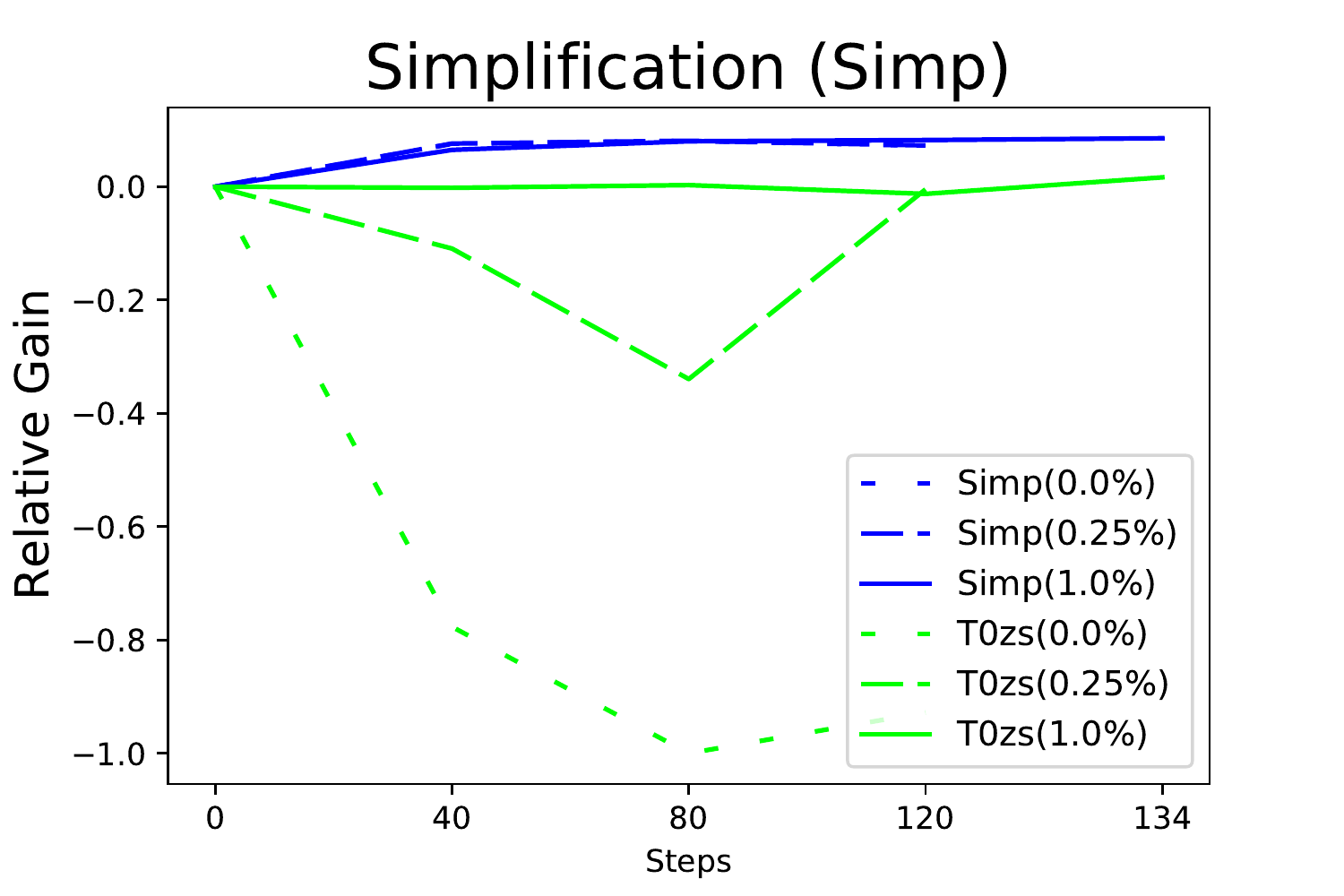}
    \includegraphics[width = 0.52\textwidth]{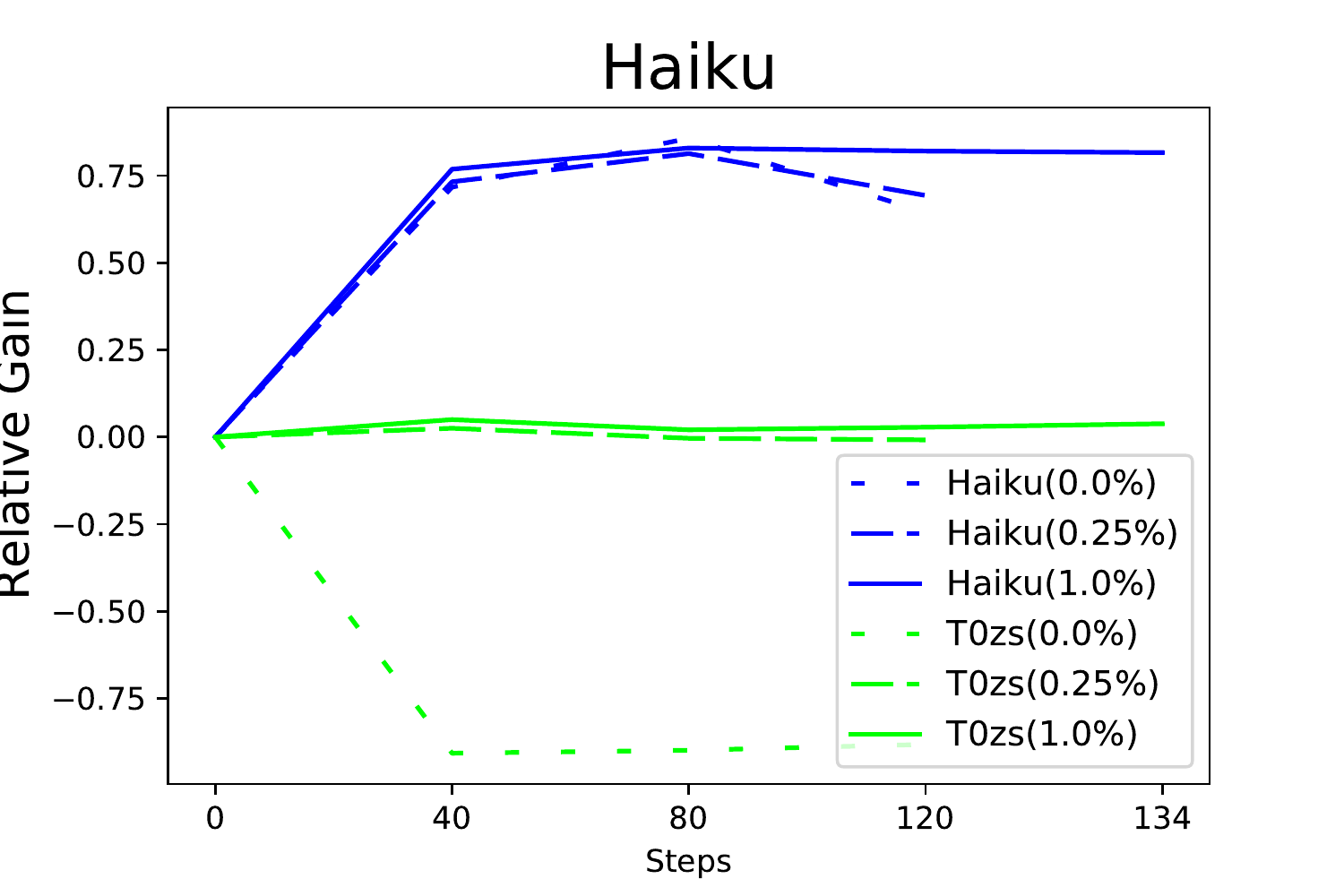}
    \caption{Rehearsal ablation with 0.0, 0.25 and 1.0\% of training data showing target task performance along with T0 zero-shot performance(T0zs) with Relative Gain in Y axis vs Number of training steps in X axis. The results are normalised in \% such that -1 corresponds to 100\% decrease and +1 means +100\% increase w.r.t. the initial performance.}
    \label{fig:rehearsal_ablation}
    \vspace{-3ex}
\end{figure}

\begin{figure*}[!ht]
    \centering
    
    \includegraphics[width=0.90\textwidth,height=0.55\textwidth]{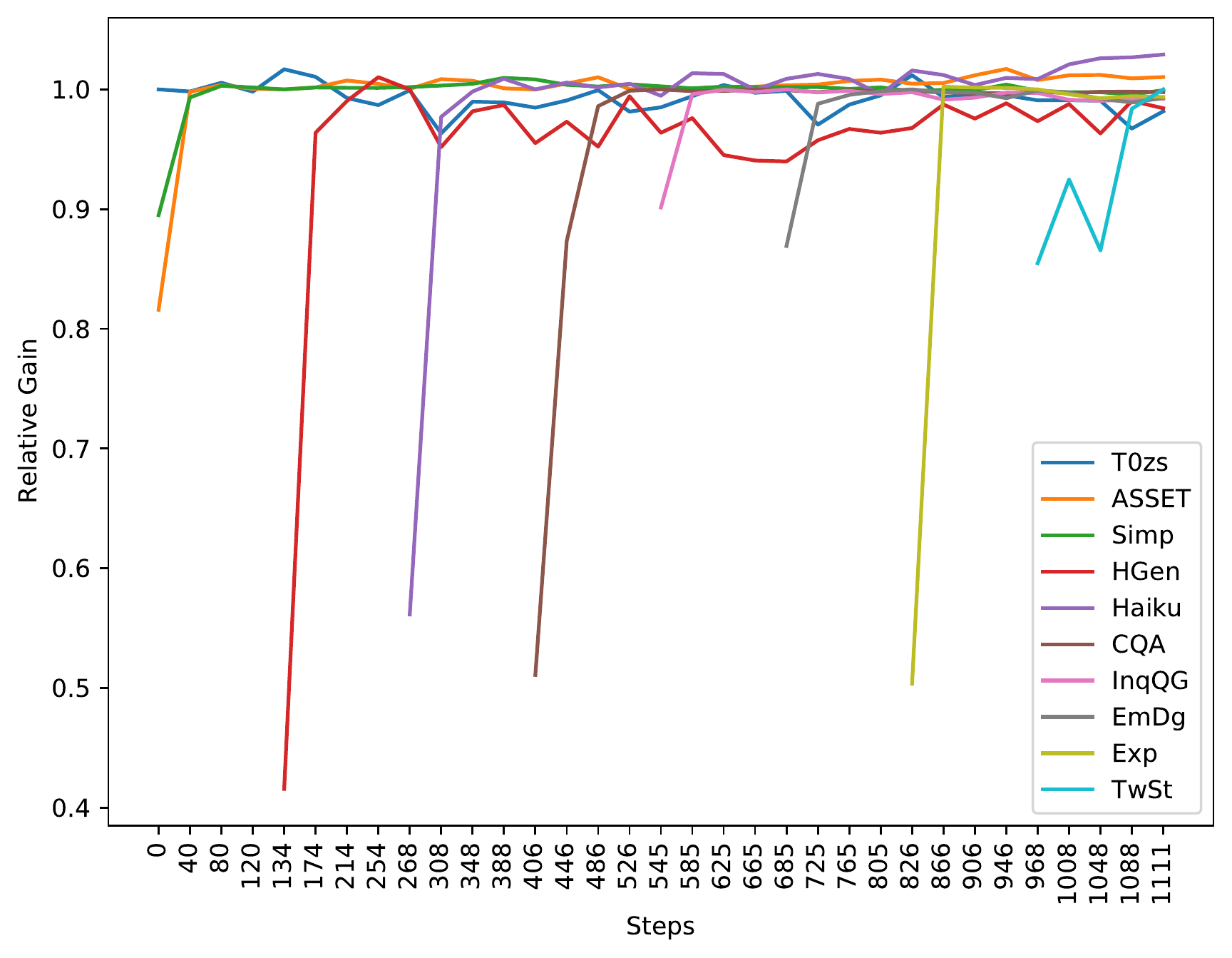}
    \vspace{-0.4cm}
    \caption{Progressive Relative Gain results for CT0 (11B) during the sequential learning(Y axis) vs Number of Training steps(X axis). The curves for tasks $T_0, ... T_7$ are displayed respectively at step $0, ..., i$ such that only the first task, Simplification (green and orange) is present at step 0, then HGen (red) etc.} 
    \label{fig:11B_seq}
    \vspace{-0.3cm}
\end{figure*}

\subsection{Learning a New Task at a time}
First, we test CLR independently on three tasks (Headline Generation with Constraint,  Simplification, and Haiku Generation), by varying the rehearsal hyper-parameter between 0\%, 0.25\% and 1\%, respectively. We report the results in terms of \textit{Relative Gain} in Figure~\ref{fig:rehearsal_ablation}.
 
We observe that for the three tasks, the rehearsal value does not affect the task result: all the blue curves are consistent. Conversely, the  rehearsal value has a dramatic impact on the T0 zero-shot results (green curves). As already discussed previously, at 0\% rehearsal,the model catastrophically forgets the T0 zero-shot tasks. Conversely, with only 0.25\% rehearsal 
we observe an almost perfect stability. Finally, with 1\% rehearsal (solid line), T0 zero-shot results are stationary, indicating that our model is able to maintain its performance on those tasks, while learning a new task.

\subsection{Learning a Sequence of New Tasks}

%SM-f to me these two paragraphs need to be before the paragraph abput Upper Bound. We want to say what we want to do, then we say we need to define some metrics/concepts to measure if it forgets or not. 
As observed from our previous experiments using Continual Learning via rehearsal we can learn a new task at any time without catastrophic forgetting, with just a very little rehearsal percentage. As a next step, we propose to measure whether language models can progressively learn more 
tasks without catastrophic forgetting. This is an important direction as it would allow the models to continually increase their knowledge and capabilities without forgetting the knowledge already acquired.

To test this hypothesis, we start from T0 checkpoint, a model trained on 50 datasets. We progressively train it on a sequence of 8 new NLG tasks (see Section \ref{section:newtask} and Table \ref{tab:task_prompt_illustration} for description of those tasks) using Continual Learning via rehearsal ($r=1\%$). We call our final model CT0. 

%SM -f I move here the first paragraph about Upper Bound. 
To measure the actual success for CL on a sequence of N tasks, we introduce the notion of \textit{Upper Bound} (UB). UB corresponds to the maximum performance achieved by the model, when fine-tuned only on a specific task, $T_n$. Arguably, the model succeeds in CL, if it maintains a performance close to UB, while learning new tasks. The normalised results, i.e .,\textit{Relative Gain} for a given task ${T_n}$, correspond to the actual scores $s$ divided by their task $T_n$ UB, $s_{T_n}/UB_{T_n}$. Hence, 1 corresponds to performing similar to the UB for any task. The model is expected to start bellow 1 before step $n$ since it has not been trained yet on $T_n$, while for the latest steps $t$ with $t > n$, results below 1 indicate task forgetting. 

In Figure \ref{fig:11B_seq}, we display the progressive sequential learning on the 8 new tasks. We learn a new task, starting from T0, and add to our rehearsal buffer 1\% of the data of the learned task. We observe an improvement progressively for each task, that is our model keeps learning new tasks. At the same time, the performance is preserved for the other tasks, (i.e., the Relative gain remains around 1) indicating the success of our CLR method in a sequential learning setup through more than 1000 gradient steps over 8 different tasks. 
%SM-f in Fig 2 not sure where each "step" is as you give the gradient steps but above you talk about "steps" w.r.t to the tasks, meaning ar step 1 is task 1, etc). So caption on Fig 2 needs clarification  

In Table \ref{tab:tab:3Bresults}, we report the results on all the 8 new tasks as well as T0tr and T0zs (see Section \ref{sec:tasks}), corresponding respectively to the evaluation sets of the 50 training datasets used in T0, and the 12 datasets kept apart for the zero-shot evaluation. In the first bloc of Table \ref{tab:tab:3Bresults}, we observe the starting performance of our two initial checkpoints, T0\_3B and T0pp(11B). The second bloc corresponds to their respective Upper Bounds. We report the results for our models after training them progressively on the 8 new tasks, as well as the baseline LAMOL (see Section \ref{sec:related_work}; for fair comparison we adapted LAMOL initialising it with T03B, additional details can be found in Appendix \ref{sec:appendix}). The CT03B and CT0pp results in Table \ref{tab:tab:3Bresults} are reported after the model was fine-tuned on the latest task in the sequence (intermediary steps are given in Table \ref{tab:appendix_detailed results} in Appendix). 
%SM-f I added this note to make clear what the number in the cells mean and to point to Table in appendix.  
%SM again, appendix has Table 7 that makes thing clear here it is not clear what the CT

Our two CT0 models obtain final results very close to their UB, maintaining 99.8\% for T0pp and 98.0\% for T0\_3B. This clearly indicates the efficiency of the CLR method. Notably, no task suffers a decrease in performance more than 2\% for T0pp. Table \ref{tab:tab:3Bresults} shows how the CT0 model remembers and retains knowledge from tasks trained at very early stages of the Continual Learning process. Moreover, CT0 still performs well on the zero-shot set of tasks (T0zs) despite no rehearsal for those. 
%SM again, appendix has Table 7 that makes thing clear here it is not clear what the CT0 cells refer to. I assume it is that peforance on that Task at the end of fine-tuning on the last task of the sequence (Twitter Stylometry) but that is not clear at all here. Needs to be explained. 

It should also be noted that the T0pp model fails to generalize for most NLG tasks, as opposed to our CT0 model. For instance Table \ref{output} in Appendix shows it can generate a haiku that has a perfect syllable count of 17 given an unseen topic of `mountain winds haunt'. It can also generate reasonable natural language explanations that often comply with our commonsense. Moreover, CT0 obtains a new state-of-the-art on the ASSET evaluation set, improving over MUSS \cite{martin2020muss}: 85.9 BLEU4 Vs 72.98 and 46.6 SARI Vs 44.15, and despite not using all the training data available.

In contrast to Continual Learning with rehearsal, LAMOL clearly diverges from its UB (T03B) indicating catastrophic forgetting. While LAMOL was known to perform well mostly on NLU tasks, we hypothesise that the generative nature for our tasks is not suited for the method. Finally, \textbf{Continual Learning with rehearsal approach is \textit{task order invariant}} as demonstrated by \textit{revfinal} results: \textit{revfinal} corresponds to CT03B trained  on the 8 tasks within in the reverse order \footnote{We report task order invariance results only using 3B and not 11B due to computing restrictions}. We give more details about the order choice in the Appendix.
%SM again here you need to say what the value is, meaning showing the performance when applying model fine-tuned on the last task in sequence (e.g ASSET in the reverse case) on that Column dataset (e.g. Haiku). 

\begin{table*}[!ht]
\renewcommand{\arraystretch}{1.5}
    \small
\begin{tabular}{|p{0.9cm}|p{0.55cm}|p{0.55cm}|p{1.25cm}|p{1.25cm}|p{1.25cm}|p{0.75cm}|p{0.65cm}|p{1.15cm}|p{0.75cm}|p{0.6cm}|l|}
\toprule
{} & T0tr &  T0zs &      ASSET &       Simp &       HGen &      Haiku &      CQA &          InqQG &      EmDg &        Exp &       TwSt \\
{} & R1 &  Acc &    B4/SARI &    B4/SARI &    R1/Cons & $H_{cust}$ &  BS &  1Tok/BS &  BS &  BS &  Clf/BS \\
\midrule
T0\_3B     & 49.8 & 48.2 &  70.1/41.0 &  12.8/41.1 &  33.6/32.2 &       34.2 &  47.6 &  2.1/58.7 &  48.6 &   32.7 &  54.4/38.0 \\
T0pp & 54.2  &  65.6 &  56.5/37.7 &  11.7/40.1 &  34.9/35.9 &       31.6 &    46.0 &  2.4/59.8 &  49.7 &   37.2 &  66.4/45.1 \\
\hline
UB\_3B  & 49.8 &  48.2   &  79.9/45.2  &   13.8/44.6    & 39.7/81.0    &    62.6       &  90.0     &    5.3/63.3  &       55.7   &    71.8      &    74.8/56.5 \\
UB\_pp  &  54.2 &  65.6 & 85.3/46.1 & 15.0/44.8 & 41.9/86.9 &   63.9    &    90.0  &   4.9/65.7    &    56.6     &  73.5     & 74.4/57.9 \\
\hline
Lamol  & 32.6 &  33.6 &  37.3/12.6 &  8.4/21.4 &  22.9/33.5 &       25.8 &  46.6 &  1.8/47.9 &  45.1 &  27.6 &  50.1/35.2 \\
CT03B  & 47.9 &  46.6 &  78.0/44.5 &  14.6/43.7 &  37.3/77.5 &       60.4 &  86.8 &  5.2/61.9 &  55.3 &  72.4 &  74.8/56.5 \\
CT0pp & \bf{53.7} & \bf{64.4} &  \bf{85.9/46.6} & \bf{14.6/44.7} &  \bf{40.7/85.5} & \bf{65.8} & \bf{89.8} & \bf{4.8/65.2} &  \bf{56.2} &  \bf{73.0} &  \bf{74.4/57.9} \\\hline\hline

revfinal   & 48.1 &  48.8 &  83.3/45.4 &  14.6/43.9 &  39.0/81.6 & 61.2 &  88.6 &  4.4/61.9 &  55.0 &   72.4 &  73.2/57.3\\

\bottomrule
\end{tabular}
    \caption{Results for the starting checkpoints T0\_3B and T0pp(11B), their upper bounds scores and our final models as well as LAMOL. Bolded result means there less than 2\% forgetting. T0tr and T0zs denote respectively for T0 train-tasks and T0 zero-shot-tasks and are the average accuracy obtained on all their evaluation datasets. B4, R1, BS denote BLEU4, ROUGE-1 and BERTScore. Note that we detail the intermediate steps results in the Appendix.}
    \label{tab:tab:3Bresults}
\end{table*}

\subsection{Zero-shot Instruction Compositionality}
\label{sec:compositionality}

Our CT0 model has learned effectively to process different instructions in specific contexts: word level constraint in the context of headline generation, or an emotional tone in the context of dialogue. Does CT0 understand these instructions in different contexts? To answer this question, and to explore whether CT0 can learn instruction compositionality we conduct several experiments.

\paragraph{Zero-Shot Constraint.} 
In Table \ref{tab:analysis_mutliple_constrain} we explore how our model succeeds in understanding constraint instructions beyond the one it was exposed during training. Our model was trained on Headline Generation with Constraint (HGen) instructions with only one match, such as \textit{Make a title for this article containing ``X''}. To test generalization, we prompt our CT0 model with unseen instructions with 2 and 3 matches, such as \textit{Make a title for this article containing ``X'' and ``Y"}, or \textit{Make a title for this article containing ``X'' and ``Y" and ``Z"}. We also compose instructions from constraint and Twitter Stylometry resulting in instructions such as \textit{Write a tweet about X, in the style of Y, containing Z}.  CT0 respects the \emph{Contain} constraint 77\% for $n=1$. The score naturally drops when $n>1$, however the satisfiablity is still 50\% of the time for $n=2$ and 40\% for $n=3$. As expected, the ROUGE-1 score also improves: \emph{NoCons}: 30.2, \#Cons=1: 38.9, \#Cons=2: 43.9 and \#Cons=3: 47.4. When we compose HGen and TwSt, CT0 also performs significantly better compared to $CT0_{No Cons}$ (46.4 vs. 10.7). 
\begin{table}[t]
\centering
\begin{tabular}{lcccc}
\toprule
{} &     \multicolumn{3}{c}{HGen} &  TwSt \\
{\# Cons} &     1 &     2 &     3 &  1 \\
\midrule
CT0    &  77.0 &  56.4 &  39.5 &     46.4 \\
$CT0_{No Cons}$ &  33.6 &  15.4 &  8.1 &     10.7 \\
\bottomrule
\end{tabular}
    \caption{Table showing Constraint generalisation i.e \% of instructions completely respected, when providing constraints for unseen prompts. $CT0_{No Cons}$ corresponds to providing the same input without constrain. 
    }
    \label{tab:analysis_mutliple_constrain}
\end{table}

\paragraph{Zero-Shot Emotional Haiku.}

\begin{figure}[ht!]
    \includegraphics[width = 0.5\textwidth]{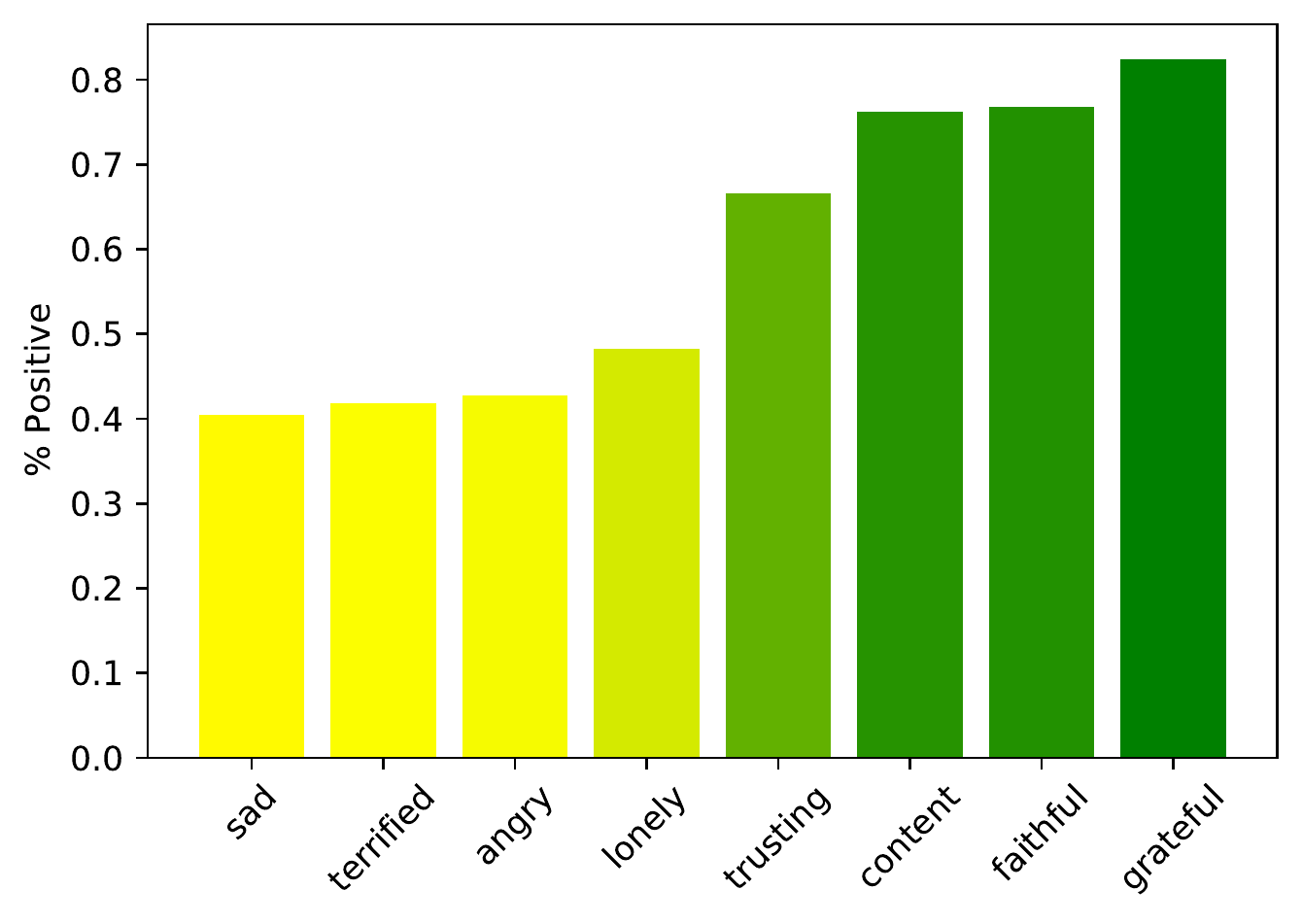}
    \caption{Emotion Generalization: Percentage of Haiku classified as positive, when adding emotion specific constraints to the Haiku instruction. We used an open source binary sentiment analysis classifier.\footnote{See on HuggingFace hub, \emph{siebert/sentiment-roberta-large-english}}}
    \label{fig:analysis_haiku_emotion}
\end{figure}

\begin{table*}[!ht]
\renewcommand{\arraystretch}{1.5}
    \small
\begin{tabular}{|p{1.4cm}|p{0.55cm}|p{0.55cm}|p{1.25cm}|p{1.25cm}|p{1.25cm}|p{0.75cm}|p{0.65cm}|p{1.15cm}|p{0.75cm}|p{0.5cm}|l|}
\toprule
{} & T0tr &  T0zs &      ASSET &       Simp &       HGen &      Haiku &      CQA &          InqQG &      EmDg &        Exp &       TwSt \\
{} & R1 &  Acc &    B4/SARI &    B4/SARI &    R1/Cons & $H_{cust}$ &  BS &  1Tok/BS &  BS &  BS &  Clf/BS \\
\midrule
UB\_rand  & N/A &  N/A & 0.5/24.3 & 0.0/29.6  & 1.5/0.1  &    9.6    &  25.2  &  1.2/25.4 & 36.3 & 33.1  & 24.7  \\

UB\_T5small  & N/A &  N/A & 87.8/45.9 & 15.6/43.2 & 35.3/67.8  & 53.4 & 54.1 & 3.4/57.0 & 51.3 & 33.8 &  52.4/54.6 \\

UB\_T53b  & N/A &  N/A & 87.0/45.6 & 15.4/43.7  & 33.0/89.4  &  63.0      & 89.9  & 2.92/61.5  & 55.3  &  71.6 & 75.6/55.4 \\
UB\_T0  & 49.8 &  48.2   &  79.9/45.2  &   13.8/44.6    & 39.7/81.0    &    62.6       &  90.0     &    5.3/63.3  &       55.7   &    71.8      &    74.8/56.5 \\
\hline
CTrand & N/A &  N/A & 0.0/22.9 & 0.0/28.5  &  0.2/0.0 &   9.6     & 25.2  & 1.2/27.9  & 28.1  &  30.7 &  24.7 \\

CT5small    & N/A &  N/A & 85.5/45.8 & 15.0/42.8 & 34.6/64.8  & 51.8 & 49.5 & 3.3/56.0 & 51.2 & 32.3 &  52.4/54.6 \\
CT53B  & N/A &  N/A &  84.6/45.8 &  14.8/44.0 & 38.3/88.3  &  62.3      & 85.8   &   4.64/62.1     &    55.5      &  73.1   & 75.6/55.4   \\
CT03B  & 47.9 &  46.6 &  78.0/44.5 &  14.6/43.7 &  37.3/77.5 &       60.4 &  86.8 &  5.2/61.9 &  55.3 &  72.4 &  74.8/56.5 \\
\bottomrule
\end{tabular}
    \caption{Results including T5-small and T5-3B, T0\_3B, and a 3B Transformer randomly initialised. We can observe that 1) only CTrand largely degrades w.r.t. its UB, UB\_rand; 2) even T5\_small is able to mostly maintain its performance indicating that scale is not what matter the most.}
    \label{tab:appendix_t5}
    \vspace{-0.3cm}
\end{table*}

We explore whether combining an emotion with the Haiku instructions would help control the haiku generation. Note that during training, only the task of Empathetic Dialogue has been exposed to emotion. Our results, reported in Figure \ref{fig:analysis_haiku_emotion},  indicate that CT0 is able to combine an emotion with the Haiku instructions in a zero-shot setting.
For instance, given the following new instruction \emph{Generate a haiku about ``held my hand''. The associated emotion is ``faithful''.}, our model output is ``\emph{He held my hand through thick and thin, Through sickness and health, through life and death}''.  A qualitative analysis also shows that CT0 understands subtle nuances; for instance given as input \emph{Generate a haiku about ``Seagulls crying high''. The associated emotion is ``nostalgic''.} our model output is ``\emph{Seagulls crying high, A familiar scene, from a childhood Now }''.

%SM-f I moved the below from the Zero-shot Constraint here as it refers to both zero-shot experiments
%These results demonstrate CT0's ability to comprehend instructions as well as to satisfy compositionality. 
These are promising results regarding CT0's ability to comprehend new instructions, including instruction composition. While contemporaneous work by \citet{nayak2022learning} propose a novel form of soft prompting for compositional zero-shot learning we show that a continually fine-tuned language model is able to perform the same.
\section{Discussion}
\label{sec:discussion}

\subsection{Why could LLMs be lifelong learners?}
\label{whyllm}

Given our current experimental protocol, one can draw different hypotheses: is CL a consequence emerging from the massive multi-task pre-training in T0? Or from the instruction tuning paradigm of T0? Or from the scaling law as studied by \citet{ramasesh2021effect}? To answer this research question, we applied the same CL setup starting from 1) T5-small, 2) T5-3B, and 3) a T5-3B architecture randomly initialised. Our results in Table \ref{tab:appendix_t5} show that CT\textbf{5} with 3B parameters performs similar to CT\textbf{0}3B on the 8 tasks. While CT\textbf{5}-small obtains as expected a lower average performance, it still mostly maintains great results w.r.t. its Upper Bound, indicating that CL does not emerge from scale. Conversely, when initialised randomly the model is not even able to obtain a good UB. These results draw a clear conclusions: \textbf{CL emerges from the intensive pre-training stage}. 
This confirms contemporaneous findings by \citet{cossu2022continual} and \citet{mehta2021empirical} in other setups and even modalities.
We report the detailed results for those experiments in the Appendix.

\subsection{Toward Concept Drift}
In the original CovidQA the task consists of answering a question present in a given paragraph. In this setup, one can arguably succeed into answering questions about COVID by transferring the task knowledge, even without particular domain knowledge about COVID. In our paper, we intentionally chose to not provide the context for CQA but only the question. This alternative setup corresponds to learning by heart the answer to a question. Our results in Table \ref{tab:tab:3Bresults} show that while we framed CQA as a new task to learn, our proposed setup also opens new way to tackle concept drift, by directly incorporating knowledge into a model. %SM-f this whole idea of concept drift and why it happens by changing the task formulation is still not clear. 

\subsection{Data Efficiency}
\label{subsec:dataEfficiency}
Our method based on rehearsal learning is simple yet efficient.  While the complexity in term of data storage and training is not constant (O(1)), with only 1\% of the previous training data we are able to retain model abilities. This result is still data and computationally efficient, compared to the standard approach of retraining the model from scratch on all tasks. In cases where the number of tasks to learn would grow by several order of magnitude, more sophisticated methods could be explored. We leave this for future research.

\begin{table}[!ht]
\small
\begin{tabular}{|l|l|}
\hline
Instr & \begin{tabular}[c]{@{}l@{}}\textit{\color{blue}Make a title for this article, finishing with}\\ \textit{\color{blue}"escalates":} the sri lankan government\\ announced the closure of government\\ schools with immediate effect as a military\\ campaign against tamil separatists escalated\\ in the north of the country .\end{tabular} \\ \hline
CT0    & \begin{tabular}[c]{@{}l@{}}sri lanka closes schools as war with\\ tamils \textbf{escalates}\end{tabular}                                                                                                                                                                                                                       \\ \hline
T0pp   & \begin{tabular}[c]{@{}l@{}}sri lanka closes schools as tamil \\ rebels advance\end{tabular}                                                                                                                                                                                                                           \\ \hline\hline
Instr &  \begin{tabular}[c]{@{}l@{}} \textit{\color{blue}Write a haiku about} `mountain winds \\haunt'\end{tabular}                                                                                                                                                                                            \\ \hline
CT0    & \begin{tabular}[c]{@{}l@{}}\textbf{mountain winds haunt}, the hollow of the\\ stones, voices echo there.\end{tabular}                                                                                                                                                                                                         \\ \hline
T0pp   & \begin{tabular}[c]{@{}l@{}}a lone tree in the mountains is haunted by\\ the wind\end{tabular}                                                                                                                                                                                                                         \\ \hline\hline
Instr & \begin{tabular}[c]{@{}l@{}}\textit{\color{blue}Explain why the two following sentences} \\ \textit{\color{blue}do not entail each other:} "Sentence 1: A\\ woman with a green headscarf, blue shirt \\ and a very big grin."; Sentence 2:"The \\ woman has been shot."\end{tabular}                                                                             \\ \hline
CT0    & \begin{tabular}[c]{@{}l@{}}A woman cannot be smiling if she has\\ been shot.\end{tabular}                                                                                                                                                                                                                             \\ \hline
T0pp   & No                                                                                                                                                                                                                                                                                                                    \\ \hline
\end{tabular}
\caption{\label{output}Outputs for HGen, Haiku and Exp from T0pp and our continually learned final model CT0.}
\end{table}

\section{Conclusion}
We explored for the first time Continual Learning for instruction-based models. Our results indicate that \emph{fine-tuned Language Models are efficient continual learners}: 1\% rehearsal is enough to maintain a high performance on previously learned tasks, while learning new ones. Additionally, we show that our model CT0 is able to comprehend new instructions obtained via instruction composition. The current technique to learn multiple tasks is to train a model from scratch. We hope this work paves the way toward a new paradigm where models do not have to be retrained all over again. We believe our experimental findings will contribute to the effectiveness of large language models, enabling them to progressively adapt to new concepts and acquire more and more abilities. As an analogy with Software Development, this could be seen as \emph{learning} new features. New checkpoints are like new versions of a model. In this context, Continual Learning will help toward the \emph{Call to Build Models Like We Build Open-Source Software}.\footnote{\url{https://tinyurl.com/3b7b2nrc}}

\section*{Acknowledgements}
We would like to thank the anonymous reviewers for their helpful comments. Tuhin is funded by Columbia Center of Artifical Intelligence \& Technology (CAIT) and the Amazon Science Ph.D. Fellowship). 

\section*{Limitations}
As discussed in \ref{subsec:dataEfficiency}, CL with rehearsal still requires to use a buffer of data previously seen which limits several scenarios where those data would not be available anymore. While we have done our best to select numerous and diverse tasks in this paper, it still represents a limited set. Would our results still hold given hundred or thousand tasks? In other modalities? It should also be noted that our study is limited to English-only datasets, as we started from T0 which is not multilingual in nature. Additionally while results using automatic metrics give a fair idea of task performance and measuring CL  abilities, we would like to conduct a human evaluation in near future although its expensive give the size of test data and the number of tasks

\section*{Ethics Statement}

Although we use language models trained on data collected from the Web, which have been shown to have issues with gender bias and abusive language, the inductive bias of our models should limit inadvertent negative impacts. Unlike model variants such as GPT, T5 is a conditional language model, which provides more control of the generated output. We have verified carefully that our training or evaluation data does not  contain any toxic text and it underwent manual inspection by the authors and experts. We also believe our work in continual learning is a step towards data efficiency and conservation of computing resources, as one saves training time by only using 1\% rehearsal
\bibliography{custom}

\begin{thebibliography}{45}
\expandafter\ifx\csname natexlab\endcsname\relax\def\natexlab#1{#1}\fi

\bibitem[{Alva-Manchego et~al.(2020)Alva-Manchego, Martin, Bordes, Scarton,
  Sagot, and Specia}]{alva-manchego-etal-2020-asset}
Fernando Alva-Manchego, Louis Martin, Antoine Bordes, Carolina Scarton,
  Beno{\^\i}t Sagot, and Lucia Specia. 2020.
\newblock \href {https://www.aclweb.org/anthology/2020.acl-main.424} {{ASSET}:
  {A} dataset for tuning and evaluation of sentence simplification models with
  multiple rewriting transformations}.
\newblock In \emph{Proceedings of the 58th Annual Meeting of the Association
  for Computational Linguistics}, pages 4668--4679, Online. Association for
  Computational Linguistics.

\bibitem[{Biesialska et~al.(2020)Biesialska, Biesialska, and
  Costa-juss{\`a}}]{biesialska-etal-2020-continual}
Magdalena Biesialska, Katarzyna Biesialska, and Marta~R. Costa-juss{\`a}. 2020.
\newblock \href {https://doi.org/10.18653/v1/2020.coling-main.574} {Continual
  lifelong learning in natural language processing: A survey}.
\newblock In \emph{Proceedings of the 28th International Conference on
  Computational Linguistics}, pages 6523--6541, Barcelona, Spain (Online).
  International Committee on Computational Linguistics.

\bibitem[{Brown et~al.(2020)Brown, Mann, Ryder, Subbiah, Kaplan, Dhariwal,
  Neelakantan, Shyam, Sastry, Askell et~al.}]{brown2020language}
Tom Brown, Benjamin Mann, Nick Ryder, Melanie Subbiah, Jared~D Kaplan, Prafulla
  Dhariwal, Arvind Neelakantan, Pranav Shyam, Girish Sastry, Amanda Askell,
  et~al. 2020.
\newblock Language models are few-shot learners.
\newblock \emph{Advances in neural information processing systems},
  33:1877--1901.

\bibitem[{Camburu et~al.(2018)Camburu, Rockt\"{a}schel, Lukasiewicz, and
  Blunsom}]{NEURIPS2018_4c7a167b}
Oana-Maria Camburu, Tim Rockt\"{a}schel, Thomas Lukasiewicz, and Phil Blunsom.
  2018.
\newblock \href
  {https://proceedings.neurips.cc/paper/2018/file/4c7a167bb329bd92580a99ce422d6fa6-Paper.pdf}
  {e-snli: Natural language inference with natural language explanations}.
\newblock In \emph{Advances in Neural Information Processing Systems},
  volume~31. Curran Associates, Inc.

\bibitem[{Cao et~al.(2021)Cao, Wei, Chen, and Wan}]{cao-etal-2021-continual}
Yue Cao, Hao-Ran Wei, Boxing Chen, and Xiaojun Wan. 2021.
\newblock \href {https://doi.org/10.18653/v1/2021.naacl-main.310} {Continual
  learning for neural machine translation}.
\newblock In \emph{Proceedings of the 2021 Conference of the North American
  Chapter of the Association for Computational Linguistics: Human Language
  Technologies}, pages 3964--3974, Online. Association for Computational
  Linguistics.

\bibitem[{Cossu et~al.(2022)Cossu, Tuytelaars, Carta, Passaro, Lomonaco, and
  Bacciu}]{cossu2022continual}
Andrea Cossu, Tinne Tuytelaars, Antonio Carta, Lucia Passaro, Vincenzo
  Lomonaco, and Davide Bacciu. 2022.
\newblock Continual pre-training mitigates forgetting in language and vision.
\newblock \emph{arXiv preprint arXiv:2205.09357}.

\bibitem[{de~Masson~D'Autume et~al.(2019)de~Masson~D'Autume, Ruder, Kong, and
  Yogatama}]{de2019episodic}
Cyprien de~Masson~D'Autume, Sebastian Ruder, Lingpeng Kong, and Dani Yogatama.
  2019.
\newblock Episodic memory in lifelong language learning.
\newblock \emph{Advances in Neural Information Processing Systems}, 32.

\bibitem[{Douillard et~al.(2021)Douillard, Ram{\'e}, Couairon, and
  Cord}]{douillard2021dytox}
Arthur Douillard, Alexandre Ram{\'e}, Guillaume Couairon, and Matthieu Cord.
  2021.
\newblock Dytox: Transformers for continual learning with dynamic token
  expansion.
\newblock \emph{arXiv preprint arXiv:2111.11326}.

\bibitem[{Fan et~al.(2019)Fan, Jernite, Perez, Grangier, Weston, and
  Auli}]{fan-etal-2019-eli5}
Angela Fan, Yacine Jernite, Ethan Perez, David Grangier, Jason Weston, and
  Michael Auli. 2019.
\newblock \href {https://doi.org/10.18653/v1/P19-1346} {{ELI}5: Long form
  question answering}.
\newblock In \emph{Proceedings of the 57th Annual Meeting of the Association
  for Computational Linguistics}, pages 3558--3567, Florence, Italy.
  Association for Computational Linguistics.

\bibitem[{French(1999)}]{french1999catastrophic}
Robert~M French. 1999.
\newblock Catastrophic forgetting in connectionist networks.
\newblock \emph{Trends in cognitive sciences}, 3(4):128--135.

\bibitem[{Jiang et~al.(2020)Jiang, Maddela, Lan, Zhong, and
  Xu}]{acl/JiangMLZX20}
Chao Jiang, Mounica Maddela, Wuwei Lan, Yang Zhong, and Wei Xu. 2020.
\newblock \href {https://www.aclweb.org/anthology/2020.acl-main.709/} {Neural
  {CRF} model for sentence alignment in text simplification}.
\newblock In \emph{Proceedings of the 58th Annual Meeting of the Association
  for Computational Linguistics, {ACL} 2020, Online, July 5-10, 2020}, pages
  7943--7960. Association for Computational Linguistics.

\bibitem[{Jin et~al.(2020)Jin, Du, Sadhu, Nevatia, and
  Ren}]{jin-etal-2020-visually}
Xisen Jin, Junyi Du, Arka Sadhu, Ram Nevatia, and Xiang Ren. 2020.
\newblock \href {https://doi.org/10.18653/v1/2020.emnlp-main.158} {Visually
  grounded continual learning of compositional phrases}.
\newblock In \emph{Proceedings of the 2020 Conference on Empirical Methods in
  Natural Language Processing (EMNLP)}, pages 2018--2029, Online. Association
  for Computational Linguistics.

\bibitem[{Jin et~al.(2021)Jin, Zhang, Zhu, Xiao, Li, Wei, Arnold, and
  Ren}]{jin2021lifelong}
Xisen Jin, Dejiao Zhang, Henghui Zhu, Wei Xiao, Shang-Wen Li, Xiaokai Wei,
  Andrew Arnold, and Xiang Ren. 2021.
\newblock Lifelong pretraining: Continually adapting language models to
  emerging corpora.
\newblock \emph{arXiv preprint arXiv:2110.08534}.

\bibitem[{Ke et~al.(2021)Ke, Xu, and Liu}]{ke-etal-2021-adapting}
Zixuan Ke, Hu~Xu, and Bing Liu. 2021.
\newblock \href {https://doi.org/10.18653/v1/2021.naacl-main.378} {Adapting
  {BERT} for continual learning of a sequence of aspect sentiment
  classification tasks}.
\newblock In \emph{Proceedings of the 2021 Conference of the North American
  Chapter of the Association for Computational Linguistics: Human Language
  Technologies}, pages 4746--4755, Online. Association for Computational
  Linguistics.

\bibitem[{Lin et~al.(2022)Lin, Wang, Lin, Jia, Xiao, Ren, and
  Yih}]{lin-etal-2022-continual}
Bill~Yuchen Lin, Sida Wang, Xi~Lin, Robin Jia, Lin Xiao, Xiang Ren, and Scott
  Yih. 2022.
\newblock \href {https://doi.org/10.18653/v1/2022.acl-long.223} {On continual
  model refinement in out-of-distribution data streams}.
\newblock In \emph{Proceedings of the 60th Annual Meeting of the Association
  for Computational Linguistics (Volume 1: Long Papers)}, pages 3128--3139,
  Dublin, Ireland. Association for Computational Linguistics.

\bibitem[{Lin(2004)}]{lin2004rouge}
Chin-Yew Lin. 2004.
\newblock Rouge: A package for automatic evaluation of summaries.
\newblock In \emph{Text summarization branches out}, pages 74--81.

\bibitem[{Lomonaco and Maltoni(2017)}]{lomonaco2017core50}
Vincenzo Lomonaco and Davide Maltoni. 2017.
\newblock Core50: a new dataset and benchmark for continuous object
  recognition.
\newblock In \emph{Conference on Robot Learning}, pages 17--26. PMLR.

\bibitem[{Madotto et~al.(2021)Madotto, Lin, Zhou, Moon, Crook, Liu, Yu, Cho,
  Fung, and Wang}]{madotto-etal-2021-continual}
Andrea Madotto, Zhaojiang Lin, Zhenpeng Zhou, Seungwhan Moon, Paul Crook, Bing
  Liu, Zhou Yu, Eunjoon Cho, Pascale Fung, and Zhiguang Wang. 2021.
\newblock \href {https://doi.org/10.18653/v1/2021.emnlp-main.590} {Continual
  learning in task-oriented dialogue systems}.
\newblock In \emph{Proceedings of the 2021 Conference on Empirical Methods in
  Natural Language Processing}, pages 7452--7467, Online and Punta Cana,
  Dominican Republic. Association for Computational Linguistics.

\bibitem[{Martin et~al.(2020)Martin, Fan, de~la Clergerie, Bordes, and
  Sagot}]{martin2020muss}
Louis Martin, Angela Fan, {\'E}ric de~la Clergerie, Antoine Bordes, and
  Beno{\^\i}t Sagot. 2020.
\newblock Muss: multilingual unsupervised sentence simplification by mining
  paraphrases.
\newblock \emph{arXiv preprint arXiv:2005.00352}.

\bibitem[{Mehta et~al.(2021)Mehta, Patil, Chandar, and
  Strubell}]{mehta2021empirical}
Sanket~Vaibhav Mehta, Darshan Patil, Sarath Chandar, and Emma Strubell. 2021.
\newblock An empirical investigation of the role of pre-training in lifelong
  learning.
\newblock \emph{arXiv preprint arXiv:2112.09153}.

\bibitem[{Mi et~al.(2020)Mi, Chen, Zhao, Huang, and
  Faltings}]{mi-etal-2020-continual}
Fei Mi, Liangwei Chen, Mengjie Zhao, Minlie Huang, and Boi Faltings. 2020.
\newblock \href {https://doi.org/10.18653/v1/2020.findings-emnlp.310}
  {Continual learning for natural language generation in task-oriented dialog
  systems}.
\newblock In \emph{Findings of the Association for Computational Linguistics:
  EMNLP 2020}, pages 3461--3474, Online. Association for Computational
  Linguistics.

\bibitem[{Mishra et~al.(2022)Mishra, Khashabi, Baral, and
  Hajishirzi}]{mishra2021cross}
Swaroop Mishra, Daniel Khashabi, Chitta Baral, and Hannaneh Hajishirzi. 2022.
\newblock \href {https://doi.org/10.18653/v1/2022.acl-long.244} {Cross-task
  generalization via natural language crowdsourcing instructions}.
\newblock In \emph{Proceedings of the 60th Annual Meeting of the Association
  for Computational Linguistics (Volume 1: Long Papers)}, pages 3470--3487,
  Dublin, Ireland. Association for Computational Linguistics.

\bibitem[{M{\"o}ller et~al.(2020)M{\"o}ller, Reina, Jayakumar, and
  Pietsch}]{moller-etal-2020-covid}
Timo M{\"o}ller, Anthony Reina, Raghavan Jayakumar, and Malte Pietsch. 2020.
\newblock \href {https://aclanthology.org/2020.nlpcovid19-acl.18} {{COVID-QA}:
  A question answering dataset for {COVID}-19}.
\newblock In \emph{Proceedings of the 1st Workshop on {NLP} for {COVID-19} at
  {ACL} 2020}, Online. Association for Computational Linguistics.

\bibitem[{Nayak et~al.(2022)Nayak, Yu, and Bach}]{nayak2022learning}
Nihal~V Nayak, Peilin Yu, and Stephen~H Bach. 2022.
\newblock Learning to compose soft prompts for compositional zero-shot
  learning.
\newblock \emph{arXiv preprint arXiv:2204.03574}.

\bibitem[{Papineni et~al.(2002)Papineni, Roukos, Ward, and
  Zhu}]{papineni-etal:2002:Bleu}
Kishore Papineni, Salim Roukos, Todd Ward, and Wei-Jing Zhu. 2002.
\newblock \href {https://doi.org/10.3115/1073083.1073135} {Bleu: A method for
  automatic evaluation of machine translation}.
\newblock In \emph{Proceedings of the 40th Annual Meeting on Association for
  Computational Linguistics}, ACL '02, pages 311--318, Philadelphia,
  Pennsylvania. ACL.

\bibitem[{Parisi et~al.(2019)Parisi, Kemker, Part, Kanan, and
  Wermter}]{parisi2019continual}
German~I Parisi, Ronald Kemker, Jose~L Part, Christopher Kanan, and Stefan
  Wermter. 2019.
\newblock Continual lifelong learning with neural networks: A review.
\newblock \emph{Neural Networks}, 113:54--71.

\bibitem[{Pedregosa et~al.(2011)Pedregosa, Varoquaux, Gramfort, Michel,
  Thirion, Grisel, Blondel, Prettenhofer, Weiss, Dubourg
  et~al.}]{pedregosa2011scikit}
Fabian Pedregosa, Ga{\"e}l Varoquaux, Alexandre Gramfort, Vincent Michel,
  Bertrand Thirion, Olivier Grisel, Mathieu Blondel, Peter Prettenhofer, Ron
  Weiss, Vincent Dubourg, et~al. 2011.
\newblock Scikit-learn: Machine learning in python.
\newblock \emph{the Journal of machine Learning research}, 12:2825--2830.

\bibitem[{Rae et~al.(2021)Rae, Borgeaud, Cai, Millican, Hoffmann, Song,
  Aslanides, Henderson, Ring, Young et~al.}]{rae2021scaling}
Jack~W Rae, Sebastian Borgeaud, Trevor Cai, Katie Millican, Jordan Hoffmann,
  Francis Song, John Aslanides, Sarah Henderson, Roman Ring, Susannah Young,
  et~al. 2021.
\newblock Scaling language models: Methods, analysis \& insights from training
  gopher.
\newblock \emph{arXiv preprint arXiv:2112.11446}.

\bibitem[{Raffel et~al.(2020)Raffel, Shazeer, Roberts, Lee, Narang, Matena,
  Zhou, Li, and Liu}]{raffel2020exploring}
Colin Raffel, Noam Shazeer, Adam Roberts, Katherine Lee, Sharan Narang, Michael
  Matena, Yanqi Zhou, Wei Li, and Peter~J Liu. 2020.
\newblock Exploring the limits of transfer learning with a unified text-to-text
  transformer.
\newblock \emph{Journal of Machine Learning Research}, 21:1--67.

\bibitem[{Rajpurkar et~al.(2016)Rajpurkar, Zhang, Lopyrev, and
  Liang}]{rajpurkar2016squad}
Pranav Rajpurkar, Jian Zhang, Konstantin Lopyrev, and Percy Liang. 2016.
\newblock Squad: 100,000+ questions for machine comprehension of text.
\newblock \emph{arXiv preprint arXiv:1606.05250}.

\bibitem[{Ramasesh et~al.(2021)Ramasesh, Lewkowycz, and
  Dyer}]{ramasesh2021effect}
Vinay~Venkatesh Ramasesh, Aitor Lewkowycz, and Ethan Dyer. 2021.
\newblock Effect of scale on catastrophic forgetting in neural networks.
\newblock In \emph{International Conference on Learning Representations}.

\bibitem[{Rashkin et~al.(2019)Rashkin, Smith, Li, and
  Boureau}]{rashkin-etal-2019-towards}
Hannah Rashkin, Eric~Michael Smith, Margaret Li, and Y-Lan Boureau. 2019.
\newblock \href {https://doi.org/10.18653/v1/P19-1534} {Towards empathetic
  open-domain conversation models: A new benchmark and dataset}.
\newblock In \emph{Proceedings of the 57th Annual Meeting of the Association
  for Computational Linguistics}, pages 5370--5381, Florence, Italy.
  Association for Computational Linguistics.

\bibitem[{Rebuffi et~al.(2017)Rebuffi, Kolesnikov, Sperl, and
  Lampert}]{rebuffi2017icarl}
Sylvestre-Alvise Rebuffi, Alexander Kolesnikov, Georg Sperl, and Christoph~H
  Lampert. 2017.
\newblock icarl: Incremental classifier and representation learning.
\newblock In \emph{Proceedings of the IEEE conference on Computer Vision and
  Pattern Recognition}, pages 2001--2010.

\bibitem[{Riabi et~al.(2021)Riabi, Scialom, Keraron, Sagot, Seddah, and
  Staiano}]{riabi2021synthetic}
Arij Riabi, Thomas Scialom, Rachel Keraron, Beno{\^\i}t Sagot, Djam{\'e}
  Seddah, and Jacopo Staiano. 2021.
\newblock Synthetic data augmentation for zero-shot cross-lingual question
  answering.
\newblock In \emph{Proceedings of the 2021 Conference on Empirical Methods in
  Natural Language Processing}, pages 7016--7030.

\bibitem[{Sanh et~al.(2022)Sanh, Webson, Raffel, Bach, Sutawika, Alyafeai,
  Chaffin, Stiegler, Scao, Raja et~al.}]{sanh2021multitask}
Victor Sanh, Albert Webson, Colin Raffel, Stephen~H Bach, Lintang Sutawika,
  Zaid Alyafeai, Antoine Chaffin, Arnaud Stiegler, Teven~Le Scao, Arun Raja,
  et~al. 2022.
\newblock \href {https://openreview.net/forum?id=OHYupm7QpfG} {Multitask
  prompted training enables zero-shot task generalization}.
\newblock In \emph{International Conference on Learning Representations}.

\bibitem[{Scialom and Staiano(2020)}]{scialom2020ask}
Thomas Scialom and Jacopo Staiano. 2020.
\newblock Ask to learn: A study on curiosity-driven question generation.
\newblock In \emph{Proceedings of the 28th International Conference on
  Computational Linguistics}, pages 2224--2235.

\bibitem[{Shin et~al.(2017)Shin, Lee, Kim, and Kim}]{shin2017continual}
Hanul Shin, Jung~Kwon Lee, Jaehong Kim, and Jiwon Kim. 2017.
\newblock Continual learning with deep generative replay.
\newblock \emph{Advances in neural information processing systems}, 30.

\bibitem[{Smith et~al.(2022)Smith, Patwary, Norick, LeGresley, Rajbhandari,
  Casper, Liu, Prabhumoye, Zerveas, Korthikanti et~al.}]{smith2022using}
Shaden Smith, Mostofa Patwary, Brandon Norick, Patrick LeGresley, Samyam
  Rajbhandari, Jared Casper, Zhun Liu, Shrimai Prabhumoye, George Zerveas,
  Vijay Korthikanti, et~al. 2022.
\newblock Using deepspeed and megatron to train megatron-turing nlg 530b, a
  large-scale generative language model.
\newblock \emph{arXiv preprint arXiv:2201.11990}.

\bibitem[{Srinivasan et~al.(2022)Srinivasan, Chang, Alva, Chochlakis, Rostami,
  and Thomason}]{srinivasan2022climb}
Tejas Srinivasan, Ting-Yun Chang, Leticia Leonor~Pinto Alva, Georgios
  Chochlakis, Mohammad Rostami, and Jesse Thomason. 2022.
\newblock Climb: A continual learning benchmark for vision-and-language tasks.
\newblock \emph{arXiv preprint arXiv:2206.09059}.

\bibitem[{Sun et~al.(2019)Sun, Ho, and Lee}]{sun2019lamol}
Fan-Keng Sun, Cheng-Hao Ho, and Hung-Yi Lee. 2019.
\newblock Lamol: Language modeling for lifelong language learning.
\newblock In \emph{International Conference on Learning Representations}.

\bibitem[{Tareaf(2017)}]{tareaf2017r}
Bin Tareaf. 2017.
\newblock \href
  {https://dataverse.harvard.edu/api/access/datafile/:persistentId?persistentId=doi:10.7910/DVN/JBXKFD/F4FULO}
  {R.: Tweets dataset-top 20 most followed users in twitter social platform}.
\newblock \emph{Harvard Dataverse}, 2.

\bibitem[{Wei et~al.(2022)Wei, Bosma, Zhao, Guu, Yu, Lester, Du, Dai, and
  Le}]{wei2021finetuned}
Jason Wei, Maarten Bosma, Vincent~Y Zhao, Kelvin Guu, Adams~Wei Yu, Brian
  Lester, Nan Du, Andrew~M Dai, and Quoc~V Le. 2022.
\newblock \href {https://openreview.net/forum?id=ewdeUNwmJLk} {Finetuned
  language models are zero-shot learners}.
\newblock In \emph{International Conference on Learning Representations}.

\bibitem[{Xu et~al.(2016)Xu, Napoles, Pavlick, Chen, and
  Callison-Burch}]{xu-etal-2016-optimizing}
Wei Xu, Courtney Napoles, Ellie Pavlick, Quanze Chen, and Chris Callison-Burch.
  2016.
\newblock \href {https://doi.org/10.1162/tacl_a_00107} {{Optimizing Statistical
  Machine Translation for Text Simplification}}.
\newblock \emph{Transactions of the Association for Computational Linguistics},
  4:401--415.

\bibitem[{Yin et~al.(2022)Yin, Li, and Xiong}]{yin-etal-2022-contintin}
Wenpeng Yin, Jia Li, and Caiming Xiong. 2022.
\newblock \href {https://doi.org/10.18653/v1/2022.acl-long.218}
  {{C}on{T}in{T}in: Continual learning from task instructions}.
\newblock In \emph{Proceedings of the 60th Annual Meeting of the Association
  for Computational Linguistics (Volume 1: Long Papers)}, pages 3062--3072,
  Dublin, Ireland. Association for Computational Linguistics.

\bibitem[{Zhang et~al.(2020)Zhang, Kishore, Wu, Weinberger, and
  Artzi}]{Zhang-etal:2020:bertscore}
Tianyi Zhang, Varsha Kishore, Felix Wu, Kilian~Q. Weinberger, and Yoav Artzi.
  2020.
\newblock \href {https://openreview.net/forum?id=SkeHuCVFDr} {Bertscore:
  Evaluating text generation with bert}.
\newblock In \emph{International Conference on Learning Representations}.

\end{thebibliography}
\bibliographystyle{acl_natbib}

\clearpage
\section{Appendix}
\label{sec:appendix}

\subsection{Tasks Order} \label{taskorder}
The task order has been selected 1) randomly among the three first tasks Text Simplifiction, Headline Generation with Constraint and Haiku Generation, and 2) in light of the actual success, we progressively kept adding new tasks. This setup corresponds to a realistic usage of our proposed method, where future tasks were thus unknown even for us. To assess a potential impact of the order, we also conduct an alternative experiment with our 3B model, where the order is reversed. We did not experimented further different orders due to the high computation required. 

\subsection{Tasks} \label{tasks}

In this section, we describe all the tasks $T$ used to progressively train and evaluate our model. For all the new tasks (i.e., not the T0 tasks), we also designed instructions, as illustrated in Table \ref{tab:task_prompt_illustration}.
\subsection{Automatic Metrics}

\subsubsection{New Tasks}\label{section:newtask}

All of our newly introduced tasks are language generation tasks in contrast to the T0 evaluation tasks and majority of the T0 training tasks (all except summarization). 

\paragraph{Text Simplification (Simpl)} \citet{acl/JiangMLZX20} provided WikiAuto, a set of 400,000 aligned sentences from English Wikipedia and Simple English Wikipedia as a resource to train sentence simplification systems. The test set contains 4,000 examples. In addition, we also evaluate our models on a second Text Simplification dataset, ASSET \cite{alva-manchego-etal-2020-asset}. This is a dataset dedicated for the evaluation of sentence simplification in English, providing 2,000 multiple references per example, unlike previous simplification datasets. Table \ref{prompts} shows our designed instructions for this task.
\vspace{-1ex}
\paragraph{Headline Generation with Constraint (HGen).} While writing a title for a news article, it can be very useful to add additional constraints, such as the presence of certain words. However, traditional decoding strategies like the BeamSearch often fail to achieve this goal as discussed in \ref{tab:analysis_mutliple_constrain}. Gigaword is one of T0 training dataset. Our new task consists of generating a title given a news article \emph{with additional constraints}. Towards this goal, for a given document D and an input keyword X we design the following three  instructions: [\textit{Make a title for this article, \textbf{starting with} / \textbf{ending with} / \textbf{that contains} ``X'' : D} where X is a word we want to be present in the output text at the beginning/end/anywhere, and D the source document, as illustrated in Table \ref{prompts}. To create the training data, we simply leverage the gold-reference to select the word X, such that our model is trained with consistent and plausible instructions. Gigaword contains millions of training examples. The original test set is composed of 1,951 examples, so we convert it to 3 sets of 1,951 examples for our Start/End/Contain instructions, respectively.

\paragraph{Haiku Generation (Haiku).} For the task of haiku generation, we crawl\footnote{The crawling part was done by Tuhin Chakrabarty and at Columbia.} 10,718 haikus with at least 1 up-vote from the Subreddit haiku, \footnote{\url{https://www.reddit.com/r/haiku/}} and split it in 9,742 and 974 example for the train and test sets, respectively. Table \ref{prompts} shows an example instruction for Haiku Generation about a given topic. 

\paragraph{Covid QA (CQA)}
\citet{moller-etal-2020-covid} created COVID-QA, a Question Answering dataset consisting of 2,019 question/answer pairs annotated by volunteer biomedical experts on scientific articles related to COVID-19. We consider this dataset since to the best of our knowledge, T0 has never been exposed to any COVID-19 related data. In its original version, the dataset is framed as SQuAD \cite{rajpurkar2016squad}, with triplets (context, question, answer), where the context contains the answer. Because T0 has been extensively trained on QA dataset, CovidQA in its original format simply requires domain transfer. To make the task more challenging, we propose to provide only the question as an input, now framing the task as ``learn the answer by heart'' in an encyclopedia style task. This way the task framing can be seen as a new strategy to incorporating knowledge and preventing the model from concept drift.

\paragraph{Inquisitive Question Generation (InqQG)} To foster long form question answering \newcite{fan-etal-2019-eli5} created the ELI5 dataset that comprises 270,000 English-language threads from the Reddit forum of the same name, \footnote{\url{https://www.reddit.com/r/ExplainLikeImfive/ }} where an online community provides answers to several open ended inquisitive questions. Table \ref{prompts} shows an example instruction in order to generate inquisitive questions. As opposed to standard Question Generation based on SQuAD, ELI5 enables open-ended questions, closer to human-style questions \cite{scialom2020ask}. We filtered out the Reddit threads to keep only well formed questions,\footnote{I.e,  starting in ``W'' or ``H'' and finishing with a question mark. See the code for the exact implementation, class ELI5promptFormat in data\_handler.py.} resulting in 61,710 and 1,681 examples for the training and test set, respectively.

\paragraph{Empathetic Dialogue Generation (EmDg)} \citet{rashkin-etal-2019-towards} proposed a benchmark for empathetic dialogue generation by creating a dataset of conversations grounded in emotional situations. Each example in the dataset contains an input emotion, situation in which dialogue appears and the entire conversation. We display in Table \ref{prompts} the corresponding instruction. At the example level, our training and test datasets contain 58,770 and 8,396 examples, respectively.

\paragraph{Explanation Generation (Exp).} The Stanford Natural Language Inference dataset consists of a classification task, where given a Premise(P) and an Hypothesis(H), the model has to chose between 3 options: 
entailed, contradiction or not related. \citet{NEURIPS2018_4c7a167b} extend this NLI dataset by annotating the explanations of the label in natural language. In our paper, we consider as input the Premise(P), the Hypothesis(H), and the label, and train our model to generate the explanation. The dataset is composed of 100,000 and 9,824 train and test examples, respectively.
\paragraph{Twitter Stylometry (TwSt)} \citet{tareaf2017r} extracted tweets from the top 20 most followed users in Twitter social platform, including singers such as Katy Perry or Selena Gomez, as well as the official account of Barack Obama when he was president of the USA. The style for tweets largely differs from one account to an another, e.g. @BarackObama: ``\emph{It's time to \#ActOnClimate}'' vs. @KimKardashian: ``\emph{makes me want to go back blonde but i'm scared it will ruin my hair :-(}''. We define the Stylometry task as generating a relevant tweet given i) a hashtag, and ii) the tweet's author. We thus selected only tweets containing hashtags (\#) from the original dataset, resulting in a total of 13,041 and 250 examples for train and test sets, respectively. We display at the bottom of Table \ref{tab:task_prompt_illustration} an example instruction for this task.

\subsection{Automatic Metrics} \label{AutoNLG}

T0 zero-shot evaluation set (see Section \ref{sec:tasks}) only contains tasks framed as classification. For T0 evaluation, \citet{sanh2021multitask} compute the loglikelihood of each of the target options, and the option with the highest log-likelihood is selected as the prediction. This strategy holds when restricting the evaluation to classification tasks. However, in the context of an open-ended model able to perform NLG tasks, a user is interested in the actual output of the model rather than probabilities. We therefore report the accuracy of the prediction compared to the ground-truth answer for all those tasks. This measure is more conservative, as it requires an exact match. 

In the context of Continual Learning, we also suspect that using only a comparison of the loglikelihood of respective classes would not reflect the actual model's memory, since the decoders are known to suffer from catastrophic forgetting more than the encoders \cite{riabi2021synthetic}. 

\paragraph{Standard NLG Metrics.} For the standard tasks, we rely on widely used metrics: ROUGE \cite{lin2004rouge} for Summarization; BLEU \cite{papineni-etal:2002:Bleu} and SARI \cite{xu-etal-2016-optimizing} for Simplification. In this paper, we also include open-domain NLG tasks, such as Dialogue or Explanation generation. The space of possible correct outputs is too large in this case to rely on n-gram based metrics like BLEU or ROUGE. For this reason, we report BERTScore \cite{Zhang-etal:2020:bertscore} to measure the similarity between a prediction and its gold-reference in those tasks.\footnote{We used BERTScore based on \textit{deberta-mnli} that is shown to have high correlation with human judgements.} 

When possible, we also designed customized metrics that are better suited for the task.\footnote{All those metrics implementations are available in the publicly released code.}
\paragraph{Customized NLG Metrics.}
\begin{itemize}[noitemsep, leftmargin=*]
%\vspace{-1.5ex}
    \item \emph{Constraint}: For our prompts with \emph{constraint}, such as ``Write a text that \emph{starts/contains/ends} with [some word]'', we also report the accuracy of respecting the constraint. Concretely, an output is correct only if it contains the [word] at the right location: the beginning for \emph{start}, the end for \emph{end}; any location for \emph{contain}.
    
    \item \emph{First Word Distribution (1Tok).} In ELI5, the questions are supposed to be inquisitive, not factual like in SQuAD. Therefore, the distribution of the first words is very informative. For instance, the percentage of questions starting with ``why/how'' is more important than ``what''. We therefore rely on the Jensen Shannon Divergence between the first words distributions of the ground truth examples and our predictions. We report its inverse, so the higher the better. 
    
    \item \emph{Author Classification (Clf)} In Twitter Stylometry, the author is part of the input, so the generated tweet is aligned with the author's style. To measure this condition, we train a classifier on the dataset, with the tweets as inputs, and the corresponding author names as target categories. We trained a Ridge Classifier using scikit-learn \cite{pedregosa2011scikit}, and obtained 0.81\% accuracy. This high accuracy allows this Clf metric to be informative enough.
    
    \item \emph{$H_{cust}$} Haiku is a type of short form poetry originally from Japan as illustrated in the Table 2. In general, it contains only 17 syllables, broken up into three lines. We calculate two differences between the prediction and the ground-truth: i) for the number of lines, and ii) for the number of syllables. \emph{$H_{cust}$} corresponds to the average of these two differences, BLEU and the Constraint satisfiability (i.e., if the generated haiku contains the topic phrase X that was present in the instruction).

\end{itemize}

\subsection{Evaluation for T0 Train Set}

Because there are 50 datasets with thousands of examples in the test sets per task, evaluating on each examples would be computationally intensive. For this reason we restricted this set to 1000 examples randomly sampled from all the examples in the test sets. Because the set contains both NLG and NLU tasks, using the accuracy is not enough. For simplicity we used therefore ROUGE-1 which allows is consistent with accuracy for NLU tasks but also allows to take into account NLG evaluation, 

\subsection{Additional Results}
\label{onetask}

In the main paper, Table \ref{tab:appendix_t5} we reported the additional results when starting from T5 and a random transformer. These results are discussed in the first section of our Discussion. 

In Table \ref{tab:appendix_detailed results} we report the progressive results, and not just the initial checkpoint, the Upper Bound and the final model. 

\begin{table*}[!ht]
\renewcommand{\arraystretch}{1.5}
    \small
\begin{tabular}{|p{1.85cm}|p{0.55cm}|p{1.25cm}|p{1.25cm}|p{1.25cm}|p{0.75cm}|p{0.65cm}|p{1.15cm}|p{0.75cm}|p{0.65cm}|l|}
\toprule
{} &  T0zs &      ASSET &       Simp &       HGen &      Haiku &      CQA &          InqQG &      EmDg &        Exp &       TwSt \\
{} &   Acc &    B4/SARI &    B4/SARI &    R1/Cons & $H_{cust}$ &  BS &  1Tok/BS &  BS &  BS &  Clf/BS \\
\midrule
T0\_3B     &  48.2 &  70.1/41.0 &  12.8/41.1 &  33.6/32.2 &       34.2 &  47.6 &  2.1/58.7 &  48.6 &   32.7 &  54.4/38.0 \\
T0pp (11B)    &  65.6 &  56.5/37.7 &  11.7/40.1 &  34.9/35.9 &       31.6 &    46.0 &  2.4/59.8 &  49.7 &   37.2 &  66.4/45.1 \\
\hline
+Simp 3B  &  \underline{48.9} &  79.9/\underline{45.2} &  13.8/\underline{44.6} &  30.3/31.0 &       30.9 &   43.9 &  2.0/56.1 &  40.2 &   34.9 &  50.8/42.5 \\
+Simp 11B &  \bf{66.7} &  85.3/46.1 &  15.0/44.8 &  34.9/36.1 &       33.0 &   47.2 &  2.1/59.0 &  48.1 &   39.2 &  68.8/47.6 \\ \hline
+HGen 3B  &  46.9 &  81.4/44.9 &  14.1/43.9 &  \underline{39.7/81.0} &       33.7 &   44.2 &  2.5/55.9 &  45.9 &  55.2 &  19.6/37.3 \\
+HGen 11B &  65.5 &  84.5/46.1 &  \textbf{15.3}/44.8 &  \bf{41.9}/86.9 &       35.9 &   46.6 &  2.9/59.7 &  48.9 &   36.4 &  69.6/48.1 \\\hline

+Haiku 3B &  48.8 &  \underline{81.6}/45.0 &  14.6/43.9 &  39.0/78.2 &       62.6 &   43.0 &  2.3/54.9 &  47.2 &   39.0 &  65.6/44.5 \\
+Haiku 11B &  64.6 &  83.5/46.1 &  14.9/45.1 &  41.1/83.0 &       63.9 &   46.0 &  2.9/59.9 &  48.9 &   37.5 &  66.4/46.2 \\ \hline
+CQA 3B &  48.5 &  79.7/44.4 &  14.0/43.8 &  37.6/75.4 &       62.2 &  \underline{90.0} &  2.0/54.4 &  42.5 &   38.7 &  66.4/45.3 \\
+CQA 11B &  64.6 &  84.3/46.1 &  14.5/\bf{44.9} &  40.9/83.7 &       63.6 &  \bf{90.0} &  2.9/59.2 &  48.5 &   42.7 &  67.2/47.3 \\ \hline
+InqQG 3B &  47.4 &  65.2/41.2 &  14.6/43.8 &  37.9/77.7 &       60.4 &  89.6 &  \underline{5.3/63.3} & 46.8 &   34.2 &  59.2/45.4 \\
+InqQG 11B &  65.5 &  85.5/46.3 &  14.9/44.8 &  40.6/81.7 &       64.5 &  89.9 &  4.9/\bf{65.7} &  49.2 &  47.7 &  61.2/45.9 \\ \hline
+EmDg 3B  &  48.6 &  73.9/43.8 &  \underline{15.0}/43.7 &  38.0/77.7 &       \underline{62.9} &  88.6 &  4.7/62.7 &  \underline{55.7} &   35.2 &  53.6/42.7 \\
+EmDg 11B  &  66.4 &  85.3/46.3 &  15.1/44.7 &  40.9/84.1 &       65.0 &  89.9 &  \textbf{5.3}/65.5 &  \bf{56.6} &   37.0 &  61.6/45.8 \\\hline
+Exp 3B   &  47.4 &  74.6/44.0 &  14.2/43.5 &  37.9/80.9 &       60.9 &  86.5 &  4.9/62.3 &  55.2 &  71.8 &  54.8/43.4 \\
+Exp 11B    &  65.0 &  85.6/46.5 &  14.9/44.7 &  40.7/84.6 &       64.5 &  89.8 &  4.8/65.5 &  56.5 &  \bf{73.5} &  63.6/46.3 \\\hline

+TwSt 3B  &  46.6 &  78.0/44.5 &  14.6/43.7 &  37.3/77.5 &       60.4 &  86.8 &  5.2/61.9 &  55.3 &  \underline{72.4} &  \underline{74.8/56.5} \\
+TwSt 11B  &  64.4 &  \bf{85.9}/46.6 &  14.6/44.7 &  40.7/85.5 &       \bf{65.8} &  89.8 &  4.8/65.2 &  56.2 &  73.0 &  \textbf{74.4}/\textbf{57.9} \\\hline\hline

rev\_final     &  48.8 &  83.3/45.4 &  14.6/43.9 &  39.0/81.6 & 61.2 &  88.6 &  4.4/61.9 &  55.0 &   72.4 &  73.2/57.3\\

\bottomrule
\end{tabular}
    \caption{Progressive results for T0 3B and 11B results for continual training set up with best 3B results underlined \& best 11B results bolded. T0zs denotes T0 zero-shot and is the average accuracy obtained on 12 eval datasets. B4, R1, BS denote BLEU-4, ROUGE-1 and BERTScore.}
    \label{tab:appendix_detailed results}
\end{table*}

\subsection{Implementation Details} \label{impl}
For all our experiments woth T0\_3B and T0pp, we instantiate our model with the T0 model \cite{sanh2021multitask} using the official implementation. \footnote{\url{https://huggingface.co/bigscience/T0pp}}

For fine-tuning T0\_3B, we used the same hyper-parameters as the ones reported in \citet{sanh2021multitask}: all the details from the batch-size to the learning rate are provided in details here. \footnote{\url{https://huggingface.co/bigscience/T0pp}} 

The only new hyper-parameter introduced in our paper is the \emph{rehearsal proportion} $r$. We explored $r \in [0, 0.25\%, 1\%]$ as reported in our first set of results.

For each task, we consider 100,000 examples for training, such that 1\% rehearsal corresponds to 1,000 examples from the memory buffer. Thus, for datasets with fewer training examples, we upsample them and conversely for largest datasets like Gigaword or Simplification, we limit to 100,000 examples.

When we scaled our best setup to the 11B parameters version of T0, \emph{T0pp},  we observed instability in validation performance. Thus, we changed the learning rate from 1e-3 to 1e-4 as well as the optimizer to AdamW instead of Adafactor for all our 11B experiments. All the other hyper-parameters remain similar to the 3B model.

For the T5 ablations, we again used the Hugging Face implementations \footnote{\url{https://huggingface.co/t5-3b}} and applied the same hyper-parameters as above. 

At inference time, we use greedy decoding, i.e. a Beam Search with $K=1$.

\end{document}